\documentclass[10pt,twocolumn,letterpaper]{article}

\usepackage[pagenumbers]{cvpr}              % To produce the CAMERA-READY version
\usepackage{graphicx}
\usepackage{amsmath}
\usepackage{amssymb}
\usepackage{booktabs}
\usepackage{pifont}% http://ctan.org/pkg/pifont
\newcommand{\xmark}{\ding{55}}%
\usepackage{subcaption}
\usepackage{multirow}
\usepackage{float}
\usepackage[flushleft]{threeparttable}
\usepackage[pagebackref,breaklinks,colorlinks]{hyperref}

\usepackage{microtype}

\usepackage[capitalize]{cleveref}
\crefname{section}{Sec.}{Secs.}
\Crefname{section}{Section}{Sections}
\Crefname{table}{Table}{Tables}
\crefname{table}{Tab.}{Tabs.}

\def\cvprPaperID{7} % *** Enter the CVPR Paper ID here
\def\confName{CVPR}
\def\confYear{2024}

\begin{document}

\newcommand{\todo}[1]{{\it\color{red} #1}}

\newcommand{\STAB}[1]{\begin{tabular}{@{}c@{}}#1\end{tabular}}

% \title{OmniHorizon: A Synthetic Omnidirectional Dataset For Outdoors Depth And Normal Estimation}
% \title{OmniHorizon: In-the-Wild Outdoors Depth and Normal Estimation \\from Synthetic Omnidirectional Dataset}
\title{Cross-Domain Synthetic-to-Real In-the-Wild Depth and Normal Estimation for 3D Scene Understanding}
% \title{Supplementary Material: Cross-Domain Synthetic-to-Real In-the-Wild Depth and Normal Estimation for 3D Scene Understanding}

\newcommand*{\affaddr}[1]{#1}
\newcommand*{\affmark}[1][*]{\textsuperscript{#1}}
\newcommand*{\email}[1]{\text{#1}}
\author{
\small Jay Bhanushali\affmark[1] \hspace{2pt} Manivannan Muniyandi\affmark[1] \hspace{2pt} Praneeth Chakravarthula\affmark[2]  \\
\affaddr{\small \affmark[1]Indian Institute of Technology Madras} \\
\affaddr{\small \affmark[2]UNC Chapel Hill}\\
}

% \author[1]{\small Jay Bhanushali}
% \author[2]{\small Praneeth Chakravarthula}
% \author[1]{\small Manivannan Muniyandi}

% \affil[1]{\footnotesize Touchlab, Indian Institute of Technology Madras, Chennai, TN- 600036, India}
% \affil[2]{\footnotesize Princeton University, Princeton, NJ 08544, USA}
% \author{\small Jay Bhanushali\\
% \small Indian Institute of Technology Madras\\
% % Chennai, India - 600036\\
% % {\tt\small jaydb99@gmail.com}
% \and
% \small Praneeth Chakravarthula\\
% \small Princeton University\\
% % First line of institution2 address\\
% % {\tt\small secondauthor@i2.org}
% \and
% \small Manivannan Muniyandi\\
% \small Indian Institute of Technology Madras\\
% % Chennai, India - 600036\\
% % {\tt\small mani@iitm.ac.in}
% }

\maketitle

\begin{abstract}
We present a cross-domain inference technique that learns from synthetic data to estimate depth and normals for in-the-wild omnidirectional 3D scenes encountered in real-world uncontrolled settings.
To this end, we introduce UBotNet, an architecture that combines UNet and Bottleneck Transformer elements to predict consistent scene normals and depth.
We also introduce the OmniHorizon synthetic dataset containing 24,335 omnidirectional images that represent a wide variety of outdoor environments, including buildings, streets, and diverse vegetation. This dataset is generated from expansive, lifelike virtual spaces and encompasses dynamic scene elements, such as changing lighting conditions, different times of day, pedestrians, and vehicles. 
Our experiments show that UBotNet achieves significantly improved accuracy in depth estimation and normal estimation compared to existing models. 
Lastly, we validate cross-domain synthetic-to-real depth and normal estimation on real outdoor images using UBotNet trained solely on our synthetic OmniHorizon dataset, demonstrating the potential of both the synthetic dataset and the proposed network for real-world scene understanding applications. 
% We plan to release the dataset, code, and trained models as open-source resources.
The dataset and accompanying code are available at \href{www.omnihorizon.github.io}{omnihorizon.github.io}.

% We introduce OmniHorizon, a synthetic dataset containing 24,335 omnidirectional images that represent a wide variety of outdoor environments, including buildings, streets, and diverse vegetation. This dataset is generated from expansive, lifelike virtual spaces and encompasses dynamic scene elements, such as changing lighting conditions, different times of day, pedestrians, and vehicles. We also present a cross-domain inference technique that learns from synthetic data to estimate depth and normals for 3D scenes encountered in real-world, uncontrolled settings. For this purpose, we introduce UBotNet, an architecture that combines UNet and Bottleneck Transformer elements to predict consistent scene normals and depth. Our experiments show that UBotNet achieves significantly improved accuracy in depth estimation (4.6\%) and normal estimation (5.75\%) compared to existing models. Lastly, we demonstrate the application of depth and normal estimation on real outdoor images using UBotNet, which has been trained solely on our synthetic OmniHorizon dataset. This demonstrates the potential of both the dataset and the network for real-world scene understanding. We plan to release the dataset, code, and trained models as open-source resources.

\end{abstract}
%  To this end, we propose UBotNet, an architecture based on a UNet and a Bottleneck Transformer, to estimate scene-consistent normals. We show that UBotNet achieves significantly improved depth and normal estimation, and have only 30\% of network parameters, compared to several exsiting networks such as U-Net with skip-connections.
\begin{figure}
    \centering
    \captionsetup{type=figure}
    \includegraphics[width=0.42\textwidth]{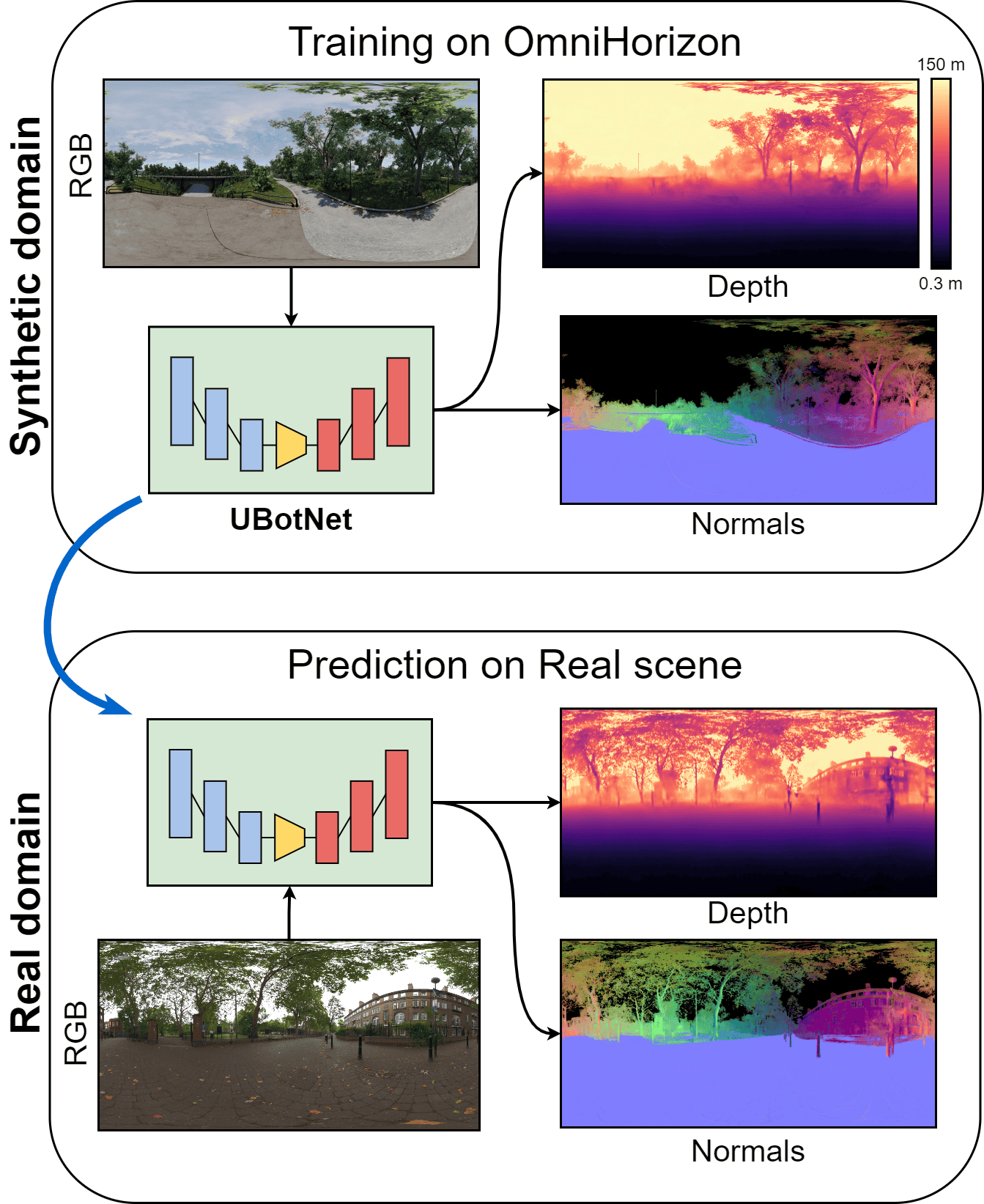}
    \captionof{figure}{\emph{Synthetic to Real cross-domain inference.} The proposed synthetic OmniHorizon dataset and the UBotNet performs cross-domain inference of scene-consistent depth and normals on real-world images captured outdoors in-the-wild.}
    \label{fig:cross-domain inference}
\vspace{-6mm}
\end{figure}

\begin{figure*}[!ht]
  \centering
  \includegraphics[width= 0.97\textwidth]{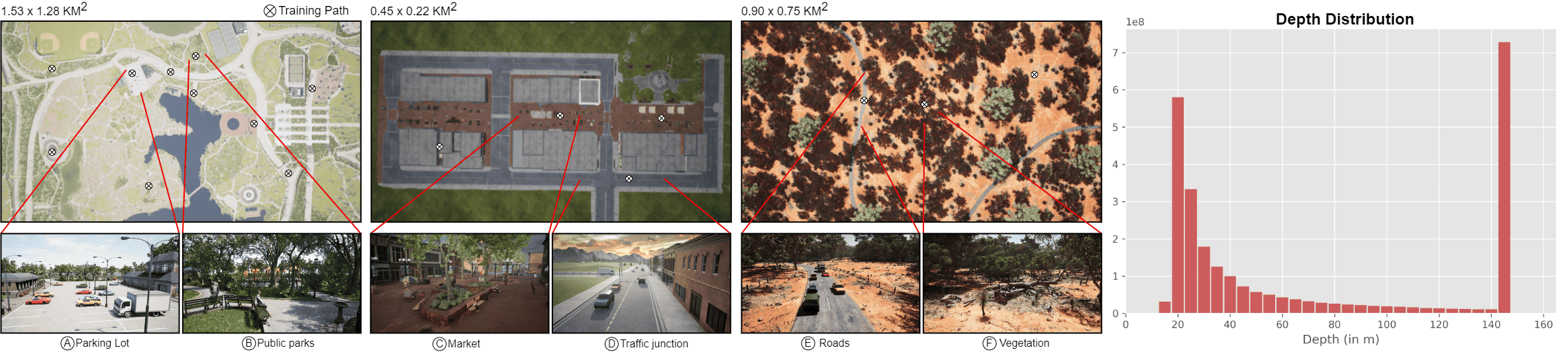}
  \caption{\emph{Overview of the OmniHorizon dataset.} Our dataset models urban areas, vegetation and various outdoor components with pedestrians and vehicles with varied depth distribution across the scenes as visualized.}
    % \caption{\emph{Overview of the OmniHorizon dataset.} Snapshot of the outdoor spaces in the dataset: City Park, Downtown and Desert. Our dataset models urban areas, vegetation and various outdoor components with pedestrians and vehicles with varied depth distribution across the scenes as visualized.}
  \label{fig:dataset}
  \vspace{-2mm}
\end{figure*}

\begin{table*}[!ht]
\centering
\caption{\emph{Comparison between the various proposed omnidirectional datasets.} While existing datasets are predominantly indoors, our proposed dataset models various outdoor environments and dynamic scene participants.}
\resizebox{0.82\textwidth}{!}{%
\begin{tabular}{@{}llllll@{}}
\toprule
 \multicolumn{1}{c}{\textbf{Dataset}} & \multicolumn{1}{c}{Domain} & \multicolumn{1}{c}{Type} & \multicolumn{1}{c} {No. of panoramic views} & \multicolumn{1}{c} {Scene Lighting} & \multicolumn{1}{c} {Dynamic components} \\ 
 
 \midrule

\multicolumn{1}{c}{Matterport3D 360$^{\circ}$ \cite{reyarea2021360monodepth}} & \multicolumn{1}{c} {Real} & \multicolumn{1}{c} {Indoors} & \multicolumn{1}{c}{9,684} & \multicolumn{1}{c}{Static} & \multicolumn{1}{c}{\xmark} \\

\multicolumn{1}{c}{Replica 360$^{\circ}$ 2k/4k RGBD \cite{reyarea2021360monodepth}} & \multicolumn{1}{c} {Real} & \multicolumn{1}{c} {Indoors} & \multicolumn{1}{c}{130} & \multicolumn{1}{c}{Static} & \multicolumn{1}{c}{\xmark} \\

\multicolumn{1}{c}{Stanford 2D-3D \cite{armeni2017joint}} & \multicolumn{1}{c} {Real} & \multicolumn{1}{c} {Indoors} & \multicolumn{1}{c}{1,413} & \multicolumn{1}{c}{Static} & \multicolumn{1}{c}{\xmark} \\

\multicolumn{1}{c}{PanoSUNCG \cite{wang2018self}} & \multicolumn{1}{c} {Synthetic} & \multicolumn{1}{c} {Indoors} & \multicolumn{1}{c}{25,000} & \multicolumn{1}{c}{Static} & \multicolumn{1}{c}{\xmark} \\

\multicolumn{1}{c}{Zillow \cite{ZInD}} & \multicolumn{1}{c} {Real} & \multicolumn{1}{c} {Indoors} & \multicolumn{1}{c}{71,474} & \multicolumn{1}{c}{Static} & \multicolumn{1}{c}{\xmark} \\

\multicolumn{1}{c}{Fukuoka \cite{mozos2019}} & \multicolumn{1}{c} {Real} & \multicolumn{1}{c} {Outdoors} & \multicolumn{1}{c}{650} & \multicolumn{1}{c}{Static} & \multicolumn{1}{c}{Vehicles \& Pedestrians} \\ 

\multicolumn{1}{c}{OmniHorizon} & \multicolumn{1}{c} {Synthetic} & \multicolumn{1}{c} {Outdoors} & \multicolumn{1}{c} {24,335}  & \multicolumn{1}{c} {Dynamic} & \multicolumn{1}{c} {Vehicles \& Pedestrians} \\ 

\bottomrule
% \multicolumn{1}{c}{} &  &  &  &  &  &  &  \\
\end{tabular}%
}
\label{table:dataset_comparision}
\end{table*}

\vspace{-3mm}
\section{Introduction}

The task of estimating depth from omnidirectional images using a single camera has received significant attention in recent years \cite{chen2021, SunSC21_HoHoNet, Pintore2021SliceNet, jiang2021unifuse, li2022omnifusion, reyarea2021360monodepth, BiFuse20}. It comes with specific challenges, such as handling distortions from equirectangular projections, and the quality and diversity of datasets are crucial for reliable depth estimation \cite{zioulis2018omnidepth}. Similarly, accurately estimating surface normals is vital for understanding scenes \cite{karakottas2019360}, especially in diverse and real-world environments \cite{chen2017surface}. In fact, previous works like GeoNet \cite{qi2018geonet} and Cross-Task Consistency \cite{zamir2020robust} have shown the benefits of jointly learning depth and surface normals. Despite a growing interest in realistic representations of real-world scenes, obtaining accurate per-pixel data from real omnidirectional images is challenging and expensive \cite{Matterport3D, Xia_2018_CVPR, mozos2019}. Existing synthetic datasets often focus on indoor spaces with limited depth range \cite{zheng2020structured3d}, making them less suitable for generalizing to outdoor scenarios with diverse scene components and larger depth ranges \cite{albanis2021, GargBR16}. While simulators like CARLA \cite{Dosovitskiy17_carla} and datasets like SYNTHIA \cite{ros2016_synthia} and Virtual KITTI \cite{cabon2020virtual} cater to autonomous driving applications, there is a notable absence of comprehensive omnidirectional datasets and robust methods for understanding scenes in various outdoor environments. This gap in research, particularly for in-the-wild monocular scene depth and normal estimation, remains significant.

In this work, we overcome these challenges and close the research gap by introducing a cross-domain synthetic-to-real neural network-based approach for depth and normal estimation for the in-the-wild 3D scene understanding. To this end, we introduce a synthetic omnidirectional dataset rendered from life-sized and diverse virtual environments, featuring randomly placed scene agents (see Figure~\ref{fig:dataset}). The primary objective is to enable the joint estimation of scene depth and normal information across various outdoor scenarios. 
% Therefore, we address limitations observed in previous omnidirectional datasets, which often a comprehensive representation of key elements. 
Subsequently, this can offer potential applications in immersive Virtual Reality \cite{Jay_2022_AIVR,Rey-Area_2022_CVPR} and Visual SLAM \cite{Won_ICRA_2020}. 
Notably, OmniHorizon dataset includes urban environments, natural elements like vegetation and rocks, and introduces dynamic elements such as pedestrians and vehicles. Additionally, the dataset encompasses different times of day, allowing for robust depth and normal estimation under varied lighting conditions.

We aim to achieve depth and normal estimation in real-world, in-the-wild scenes using a network trained exclusively on just synthetic dataset. In doing so, we also examine and address limitations in existing neural network architectures designed for depth and normal estimation. We propose an enhanced network architecture named UBotNet, drawing inspiration from U-Net \cite{RonnebergerFB15_UNet} and the Bottleneck transformer \cite{Srinivas_2021_CVPR}, which notably improves depth and normal estimation for both synthetic and real-world scenes. Furthermore, we conduct a thorough analysis of the cross-domain inference performance of UBotNet, trained on our OmniHorizon dataset, and the state-of-the-art Fukuoka dataset \cite{mozos2019}. The introduced dataset and neural network demonstrate significant advancements in cross-domain inference, showcasing the capability to train the network on synthetic scenes and successfully apply it to comprehend real-world, in-the-wild scenes, see Figure~\ref{fig:cross-domain inference}.

%We identify an important drawback with vanilla U-Net architecture when estimating normals on our dataset. We propose a new architecture, U-BotNet, based on U-Net and Bottleneck transformer to improve the normal estimation for both synthetic and real-world scenes.
%Moreover, we are also interested in understanding the generalization ability of a network trained on the synthetic dataset to estimate depth and normals for the images from the real-world domain (see Figure~\ref{fig:cross-domain inference}). To this end, we demonstrate the improvements in the performance of the network trained on our dataset, on Fukuoka dataset. 
% To summarise, our two core contributions in this work:
% To summarise, our core contributions in this work:

\vspace{1mm}
In summary, we make two key contributions:
\begin{itemize}
    % \vspace{-4pt}
    \item \textbf{OmniHorizon}: We introduce a synthetic omnidirectional dataset comprising over 24,000 images, designed for comprehensive scene depth and normal estimation. This dataset is well-suited for cross-domain inference, featuring diverse landscapes, dynamic elements such as varying lighting, cloud formations, pedestrians, and vehicles.
    \item \textbf{UBotNet}: We propose a novel network architecture, UBotNet, inspired by U-Net and the Bottleneck Transformer. UBotNet is tailored for efficient depth and consistent scene normals estimation, demonstrating generalizability for cross-domain inference. Additionally, we introduce a streamlined variant, UBotNet Lite, with 71\% fewer parameters, emphasizing compactness and efficiency in the network design.
\end{itemize}

    % \item \textbf{Dynamic scene elements}: Flexible dataset generation pipeline for introducing dynamic scene components such as dynamic lighting, pedestrians and vehicles.
\section{Related Work}
% We broadly classify the omnidirectional datasets in the literature related to depth and normal estimation into two categories based on whether the data was curated from the real world (Realistic dataset) or generated using a 3D rendering engine (Synthetic dataset). An overview of the datasets are also presented in Table~\ref{table:dataset_comparision}.

We categorize omnidirectional datasets in the literature concerning depth and normal estimation into two main groups. These are based on whether the data is collected from real-world scenarios (Real Datasets) or generated using a 3D rendering engine (Synthetic Datasets). Table~\ref{table:dataset_comparision} provides an overview of these datasets.

\vspace{-2mm}
\paragraph{Real Datasets} 
Matterport 3D \cite{Matterport3D} is a real-world dataset capturing indoor scenes, comprising of 10,800 panoramic views from 90 building-scale environments. It provides data including depth, normals, surface reconstruction, camera poses, and semantic segmentations derived from these scenes. Matterport3D 360$^{\circ}$ \cite{reyarea2021360monodepth} is an extension, adding 9,684 high-resolution 360 samples specifically designed for monocular depth estimation. Gibson \cite{Xia_2018_CVPR} offers a virtual environment based on real-world settings, delivering photorealistic interiors with RGB images, depth information, surface normals, and semantic annotations for selected spaces. 
Stanford2D3D \cite{armeni2017joint} presents a dataset gathered from six large-scale indoor areas, consisting of 1,413 equirectangular RGB images along with corresponding depths, surface normals, and additional data. HM3D \cite{hm3d} stands out as the most extensive dataset for 3D indoor spaces, providing 1.4 to 3.7 times the navigable space compared to other datasets. 
Replica \cite{straub2019replica} comprises 18 3D indoor scene reconstructions, while Replica 360$^{\circ}$ 2k/4k RGBD \cite{reyarea2021360monodepth} extends this dataset, offering 130 RGB-D pairs rendered at resolutions of $2048 \times 1024$ and $4096 \times 2048$. Zillow \cite{ZInD} is one of the largest indoor datasets, featuring 71,474 panoramas, 21,596 room layouts, and 2,564 floor plans, all captured from 1,524 homes. Fukuoka \cite{mozos2019}, designed for place categorization challenges, is an outdoor dataset, providing 650 panoramic RGB views, 3D depth, and reflectance maps. The dataset encompasses various outdoor settings such as forests, urban areas, coastal regions, parking lots, and residential areas.

\vspace{-2mm}
\paragraph{Synthetic Datasets} 
% Structured 3D \cite{zheng2020structured3d} is a synthetic indoors dataset with 3500 scenes providing multiple furniture configurations. They also provide warm and cold lighting conditions in the dataset. PanoSUNCG \cite{wang2018self} provides 103 scenes with 25K omnidirectional images rendered using environments from SUNCG \cite{song2016ssc}. 360D which was introduced in {\cite{zioulis2018omnidepth}} provides 360 color images along with corresponding depth rendered from two synthetic (SunCG, SceneNet \cite{McCormac2017}) and two realistic (Matterport 3D, Stanford2D3D) datasets.

% As compared in Table~\ref{table:dataset_comparision}, our OmniHorizon dataset contains outdoor virtual spaces with dynamic scene lighting and scene participants. This bridges a significant gap in existing datasets which predominantly considers indoor environments with static scene components and lack context for outdoor spaces.

Structured 3D \cite{zheng2020structured3d} is a synthetic indoor dataset featuring 3,500 scenes, each offering various furniture configurations. The dataset also incorporates diverse lighting conditions, including warm and cold settings. PanoSUNCG \cite{wang2018self} contributes 103 scenes, rendering 25,000 omnidirectional images using environments from SUNCG \cite{song2016ssc}. 360D, introduced by Zioulis et al.\cite{zioulis2018omnidepth}, includes 360 color images with corresponding depth, rendered from two synthetic datasets (SunCG, SceneNet \cite{McCormac2017}) and two realistic datasets (Matterport 3D, Stanford2D3D). As highlighted in Table~\ref{table:dataset_comparision}, our OmniHorizon dataset stands out by encompassing outdoor virtual spaces, complete with dynamic scene lighting and diverse scene participants. This addresses a significant gap in existing datasets, which predominantly focus on indoor environments with static scene components and lack contextual information for outdoor spaces.

\vspace{-2mm}
\paragraph{Monocular Omnidirectional Depth and Normals}
% There has been a large body of work on estimating depth and/or surface normals from a single image, all of which use dense ground truth depth or normals during training \cite{godard2017unsupervised, liu2015deep, eigen2014depth, li2015depth, eigen2015predicting, wang2015towards, wang2015designing, karsch2014depth}. 
%
% bla bla bla
%
Early approaches to monocular omnidirectional depth estimation, pioneered by \cite{tateno2018distortion} and \cite{zioulis2018omnidepth}, involved adapting traditional CNNs for spherical images, either through distortion-aware training on perspective images or by introducing a rendered spherical dataset. Notably, Pano Popups \cite{eder2019pano} concurrently predicted depth and surface orientation, emphasizing the challenges in approximating planar regions. The significance of spatially imbalanced predictions in 360° depth estimation was addressed by Generalized Mapped Convolutions \cite{eder2019mapped}, showcasing the importance of accounting for distortion in equirectangular projections. Omnidirectional extension networks \cite{cheng2020omnidirectional} introduced a near field-of-view (NFoV) perspective depth camera alongside a spherical one, enhancing detail preservation in inferred depth maps. Recent works have explored diverse paths, with approaches like BiFuse \cite{wang2020bifuse} and UniFuse \cite{jiang2021unifuse} focusing on the fusion of cubemap and equirectangular features. HoHoNet \cite{sun2021hohonet} adapted classical CNNs for 360° images, flattening meridians to DCT coefficients for efficient dense feature reconstruction in monocular depth estimation from spherical panoramas. Other studies, including \cite{jin2020geometric, zeng2020joint}, investigate the relationship between layout and depth estimation, while \cite{feng2020deep} explores joint optimization of depth and surface orientation using a UNet model. 
However, these methods encounter challenges in joint estimation of depth and normals, and generalizing to images in real-world scenarios due to the limited scene diversity inherent in existing datasets acquired by depth sensors. In addition, normals estimated using depth is considerably inaccurate in comparison to direct normal estimation \cite{Bae2022irondepth}.
% \textcolor{blue}{Additionally, the authors of IronDepth \cite{Bae2022irondepth} found that the accuracy of surface normals estimated from the output of depth estimation methods is considerably lower than the accuracy achieved when surface normals are directly estimated by neural networks. This issue holds particular significance in the context of omnidirectional images, given that existing works commonly calculate normals from point clouds generated using predicted depth maps.}
%
In contrast, our proposed hybrid neural network architecture allows for cross-domain inference and joint estimation of depth and normals, generalizing well to in-the-wild scenes.

%The common feature of the above datasets is the photo-realistic representation of the interior spaces along with additional data such as corresponding depth, surface normals and semantic labels. The lack of outdoor spaces and elements in the above datasets is one of the primary motivations of this work. Representing outdoor spaces provides it's own unique set of challenges due to vast number of outdoor entities. Likewise, sky, time of day settings and dynamic lighting are important aspects of an outdoor scenario which are often neglected in the indoor datasets. 
\vspace{-2mm}
\section{Dataset}
% We rendered the OmniHorizon dataset using Unreal Engine 4.
% As part of the dataset, we provide color images, scene depth and world normal in stereo (top-bottom) format. All data is rendered at $1024 \times 512$ resolution. All the scene assets for the dataset were acquired from the Unreal Marketplace. We designed \emph{training path} which is an animated sequence of 1521 frames captured using a moving camera. For each scene, we rendered training paths depending on the scale of the scene (see Figure \ref{fig:dataset}). As a result, the dataset contains 24,335 omnidirectional views for outdoors scene depth and normal estimation. We clamp the depth to 150m (Unreal units) and use world-space normals for the normal maps (see supplementary material for discussion). We used an image split of 85 : 15 for the training (19392) and the validation (3423) sets. Specifically, we reserved the entire training-path 4 (Figure~\ref{fig:dynamic_lighting} and Figure~\ref{fig:datasetResults}) from City Park space as test image data (1520 images). The training-path 4 is isolated from other paths in the scene and contains underpass, stairs, uneven terrain, building and pedestrians, thereby making it an ideal test image data. We discuss in detail several attributes of the dataset in the following subsections:

The OmniHorizon dataset was generated using Unreal Engine 4 \cite{unrealengine}, featuring color images, stereo scene depth, and world normals in a top-bottom format, all rendered at $1024 \times 512$ resolution. Utilizing assets from the Unreal Marketplace, we designed an animated \emph{training path} with 1521 frames captured using a moving camera for each scene (see Figure \ref{fig:dataset}), resulting in 24,335 omnidirectional views for outdoor scene depth and normal estimation. Scenes were scaled appropriately, and the dataset includes underpass, stairs, uneven terrain, buildings, and pedestrians. Depth is capped at 150m (Unreal units), and world-space normals are employed for normal maps, see supplementary material for discussion. The training-validation split is 85:15, with 85\% (20,685) for training and 15\% (3,650) for validation. 
Notably, we generate an unseen scene sequence with diverse elements, including underpasses, stairs, uneven terrain, buildings, and pedestrians, isolated from training data, to serve as a test set (1520 images). Additional attributes of the dataset are discussed in subsequent subsections.

\begin{figure*}[!ht]
  \centering
%   \begin{subfigure}{0.4\linewidth}
%     \includegraphics[width=\linewidth]{Figures/dynamic_lighting_schematic.png}
%     \caption{}
%     \label{fig:first}
% \end{subfigure}
% \hfill
% \begin{subfigure}{0.59\linewidth}
%     \includegraphics[width=\linewidth]{Figures/varying-day-of-lighting.png}
%     \caption{}
%     \label{fig:second}
% \end{subfigure}
  \includegraphics[width=.88\linewidth]{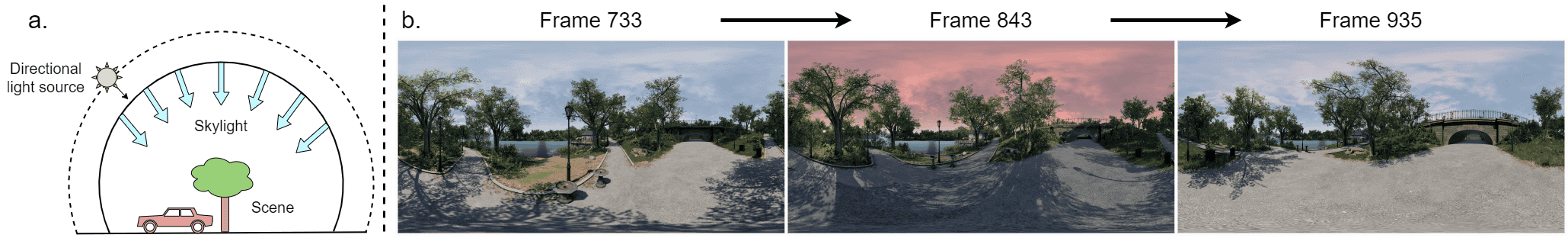}
  \vspace{-3pt}
  \caption{\emph{Dynamic lighting and varying time of day settings.} a) The lighting of the scene is varied by modulating the directional light (sun) and secondary light source (skylight). b) Changes in the scene lighting condition achieved using the modulation of the light sources.}
  \label{fig:dynamic_lighting}
  \vspace{-2mm}
\end{figure*}

\subsection{Scene Attributes}
% We notice that the scene attributes and context plays a crucial role in the performance of the neural networks on real-world scene inferences \cite{albanis2021}. Our dataset is designed to capture a wider variety of scene attributes including urban environments consisting of buildings and roads, as well as more naturally occuring uneven terrains and vegetation, and several other entities that generally make an outdoors environment. Figure~\ref{fig:dataset} provide a snapshot overview of our dataset. For example, \emph{Downtown}\cite{downtown2022} and \emph{CityPark}\cite{citypark2022} scenes of our dataset represent urban areas covering buildings, houses, parks and other street props. While CityPark scenes contain wider roads and streets, Downtown scenes are composed of market, narrower streets and alley ways.  Scene such as \emph{Desert}\cite{ruralAustralia2022} is composed of rocks, roads, uneven terrain and wild vegetation.

The performance of neural networks in real-world scene inferences is significantly influenced by scene attributes and context \cite{albanis2021}. Our dataset is designed to cover a diverse range of scene attributes, including urban environments with buildings and roads, as well as naturally occurring uneven terrains, vegetation, and various outdoor elements. Figure~\ref{fig:dataset} provides a snapshot overview of our dataset, featuring scenes like Downtown \cite{downtown2022} and CityPark \cite{citypark2022}, which represent urban areas with buildings, houses, parks, and street props. CityPark scenes include wider roads, while Downtown scenes consist of markets, narrower streets, and alleyways. Additionally, scenes like Desert \cite{ruralAustralia2022} feature rocks, roads, uneven terrain, and wild vegetation.

% \todo{This paragraph needs to rewritten. Its confusing.}
% While the networks can be trained with the remaining outdoors only scenes, we found that the normals would often lack the details that can be observed in the input image. We found that the Dungeons scene provides better close proximity context, which helps the network to capture close-up the details from the input image. We attribute this to the overall detailed indoor structures and stone walls in the Dungeons scene.

% \begin{figure}[!ht]
%   \centering
%   \includegraphics[width=.80\textwidth]{Figures/pedestriansAndCars.png}
%   \caption{Pedestrians And Traffic in the OmniHorizon Dataset.}
%   \label{fig:pedestriansAndCars}
% \end{figure}

\subsection{Dynamic Lighting}

In outdoor environments, lighting conditions vary dynamically based on the time of day and intricate cloud patterns in the sky. Existing datasets often lack the modeling of such dynamic lighting changes, leading to compromised performance in trained neural networks, particularly for in-the-wild scene understanding tasks in outdoor settings. Note that scene depth and normals remain independent of scene brightness or color. To address this, we adopted a two-pass rendering approach, isolating scene depth and normal data from scene color. This separation allowed us to prototype changes in scene lighting, brightness, and color while ensuring consistent depth and normal data generation during rendering. We simulated dynamic lighting changes by modulating the position and intensity of both a directional light source (sunlight) and a secondary light source (diffuse light from the sky) throughout an animated sequence, mimicking a full day. Figure~\ref{fig:dynamic_lighting} illustrates example changes in lighting conditions achieved by modulating these light sources. To capture more complex lighting variations resulting from different cloud formations in the sky, we utilized a sky plugin \cite{goodsky} to render various sky-cloud settings, including Stratus, Cumulus, and Cirriform clouds. Cloud coverage was varied from very light to extremely heavy, and the dataset spans early morning to late evening time settings.

% \vspace{-5mm}
\subsection{Dynamic Scene Participants}
% Of the several entities that make outdoor spaces, vehicles and pedestrians form important scene components. To modeled these dynamic scene participants, we model multiple classes of vehicles including trucks, hatchbacks, SUVs, pickup and sports cars. We rendered both automatically and randomly placed vehicles in outdoor environments  as well as manually placed vehicles in parking lots and roads (see Figure \ref{fig:dataset}).
%Vehicles and pedestrians are important components of the outdoor spaces. 
%There are multiple class of vehicles used in the dataset which includes: truck, hatchback, pickup, SUV and sports car. We have manually placed the vehicles in the parking lots and roads. 
% We further used 3D scanned avatars \cite{realisticAvatars2022} and Metahumans\cite{metahumans2022} to increase the visual diversity of the pedestrians used in the dataset.
\begin{figure}
    \centering
    \captionsetup{type=figure}
    % \vspace{-2pt}
    \includegraphics[width=0.46\textwidth]{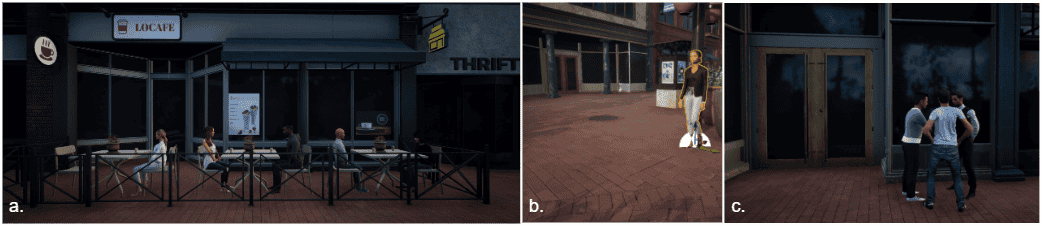}
    \captionof{figure}{\emph{Examples of pedestrians in OmniHorizon dataset.} a) virtual avatars sitting in a cafeteria, b) pedestrian walking on the street (spline path is highlighted in pink) and c) casual group hangout.}
    \label{fig:vehiclesAndPedestrians}
\vspace{-4mm}
\end{figure}
% Metahumans used in our dataset are high-fidelity realistic virtual avatars with diverse skin tones and detailed grooming. We have used highest LOD level (LOD $0$) for Metahumans (see supplementary material). For rendering human scene participants, we used three settings: idle poses, sitting and walking. The walking behaviour and trajectories of the pedestrians are controlled using spline path and the blueprint. Figure~\ref{fig:vehiclesAndPedestrians} depicts several examples of the placement of human avatars in the dataset. It can be seen that the avatars are placed throughout the scene in various realistic locations such as sitting outside a cafeteria, walking on the street and discussing in a group.

Dynamic scene elements such as vehicles and pedestrians are predominant and play crucial roles in outdoor spaces. To accurately represent these components, we modeled various classes of vehicles, including trucks, hatchbacks, SUVs, pickup trucks, and sports cars. These vehicles were randomly generated and placed in outdoor environments, as well as manually positioned in parking lots and on roads (see Figure \ref{fig:dataset}). Additionally, we introduced visual diversity in pedestrians by incorporating 3D scanned avatars \cite{realisticAvatars2022} and high-fidelity realistic Metahumans \cite{metahumans2022}. Metahumans, with diverse skin tones and detailed grooming, were utilized at the highest Level of Detail (LOD 0) to enhance visual realism (see supplementary material). The pedestrians exhibit three distinct settings: idle poses, sitting, and walking. Walking behavior and trajectories are controlled using spline paths and Unreal Engine's blueprints. Figure~\ref{fig:vehiclesAndPedestrians} showcases examples of human avatars strategically placed throughout the dataset, engaging in realistic activities such as sitting outside a cafeteria, walking on the street, and engaging in group discussions.

\begin{figure*}[!ht]
  \centering
%   \includesvg [width=\textwidth]{../Figures/UBotNet_architecture.svg}
  \includegraphics[width=0.82\textwidth]{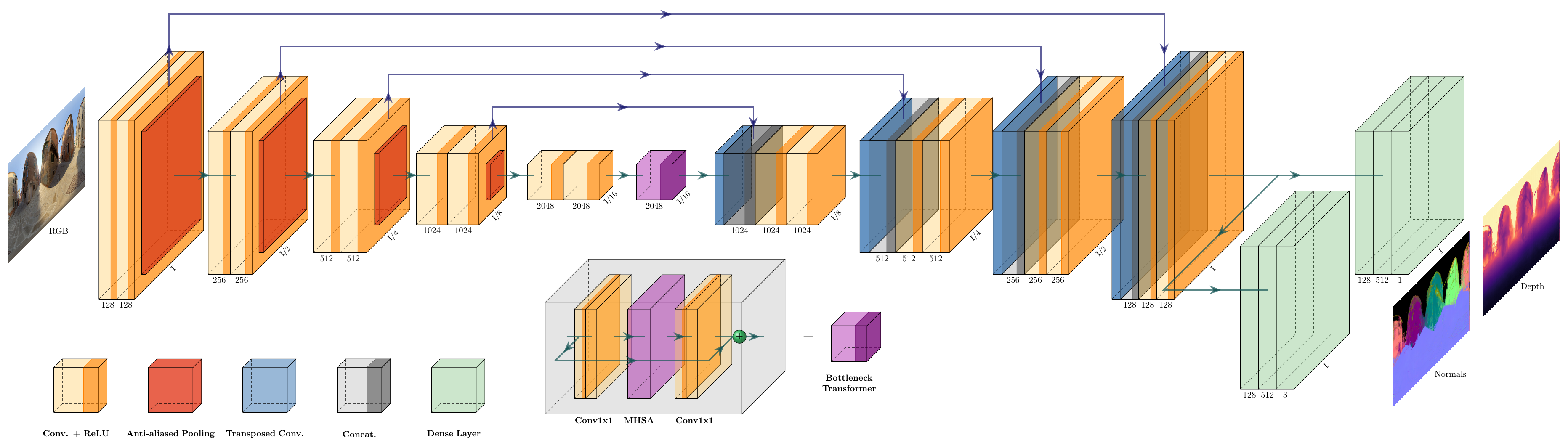}
  \caption{\emph{Proposed UBotNet architecture.}  UBotNet is a hybrid architecture based on UNet and Bottleneck Transformer (BoTNet). Anti-aliased max pooling is used for the pooling operation. The transformer block is placed in the middle of the encoder and decoder paths of the UNet. UBotNet Lite uses separable convolutions in place of standard convolution layers; otherwise, it is identical to UBotNet. A simplified illustration of BoTNet is also shown which contains Multi-Head Self-Attention (MHSA) for learning global context.}
  \label{fig:unetTransformer}
  \vspace{-3mm}
\end{figure*}

\vspace{-2mm}
\section{Neural Cross-domain Inference}
% In this section, we describe UBotNet architecture based on U-Net\cite{RonnebergerFB15_UNet} and a Bottleneck transformer\cite{Srinivas_2021_CVPR} for cross-domain inference, and our network training procedure.
% We conduct four different experiments for evaluating the proposed dataset and network architectures: a) Benchmark on OmniHorizon dataset, b) Ablation study of the dataset, c) Sim-to-Real domain transfer, and d) In-the-wild depth and normal estimation from real-world omnidirectional images images.

In this section, we present the UBotNet architecture, inspired from U-Net \cite{RonnebergerFB15_UNet} and Bottleneck transformer \cite{Srinivas_2021_CVPR}, for cross-domain inference, along with our network training methodology. Our evaluation comprises four distinct experiments: a) Benchmarking on the OmniHorizon dataset, b) Ablation study of the dataset, c) Sim-to-Real domain transfer performance, and d) In-the-wild depth and normal estimation from real-world omnidirectional images.

\subsection{UBotNet Architecture}

In comparison to high-capacity encoders like ResNet and DenseNet architectures, UNet with skip-connections has exhibited superior performance in Pano3D benchmarks \cite{albanis2021}. However, U-Net architecture is unsuitable for predicting consistent normals across both synthetic and real-world scenes (see Figure \ref{fig:datasetResults} and additional results in supplementary material). To address this limitation and enable the network to capture information in a broader context with long-range dependencies, we sought inspiration from Vision Transformers (ViT). ViT, known for achieving state-of-the-art results in image classification using a pure transformer architecture \cite{Dosovitskiy2020}, has demonstrated a wider receptive field compared to CNNs, allowing global integration of information across an image. Recent studies replacing the final layers of ResNet with a bottleneck transformer have shown improved performance in instance segmentation and object detection tasks \cite{Srinivas_2021_CVPR}. The fusion of U-Net with attention or transformer-based architectures has also been explored, particularly in medical image segmentation \cite{chen2022_transunet,otkay2018_attentionUNet}.

Building upon these insights from prior research, we introduce an enhanced architecture named \emph{UBotNet}, that can efficiently learn local features through convolutional layers and integrate self-attention for global context aggregation. UBotNet combines elements from U-Net and the Bottleneck transformer. Specifically, we position the self-attention transformer block at the lowest resolution feature maps in the U-Net bottleneck, as self-attention involves $O(n^2d)$ memory and compute \cite{vaswani2017attention}. Figure~\ref{fig:unetTransformer} provides an overview of our proposed architecture. To enhance its performance, we replace the traditional max-pooling layer with an anti-aliased max-pooling layer {\cite{zhang2019shiftinvar}}. Additionally, we present a streamlined compact version of UBotNet, dubbed \emph{UBotNet Lite}, employing separable convolution {\cite{Chollet_2017_CVPR}} to significantly reduce the number of parameters. UBotNet Lite (38.3M) has a 71.2\% reduction in parameters compared to its larger counterpart, UBotNet (133M). Towards the end, we incorporate two branches of fully-connected layers with sigmoid activation to predict scene depth and consistent normals. The CNN blocks focus on capturing local image features, while the Multi-Head Self-Attention (MHSA) block from the bottleneck transformer learns global contextual features. See supplementary material for additional details. 
We demonstrate and validate the impact of learning local and global-scale features in the experiments detailed in Section~\ref{sec:discussion}.
% Demonstrating the importance of learning both local and global-scale features, we validate and showcase their impact in the experiments detailed in Section~\ref{sec:discussion}.
\begin{table*}[!ht]
\centering
\caption{\emph{Quantitative Results for the benchmark evaluated on the OmniHorizon dataset.} Values in \textbf{bold} highlight best results. (* denotes networks that only perform depth estimation)}
% \todo{please use a different color. red gives negative vibes}
\resizebox{0.92\textwidth}{!}{%
\begin{tabular}{@{}lllllccclllccc@{}}
\toprule
\multicolumn{1}{c}{} & \multicolumn{1}{c}{} & \multicolumn{3}{c}{\textbf{Depth Error $\downarrow$}} & \multicolumn{3}{c}{\textbf{Depth Accuracy $\uparrow$}} & \multicolumn{3}{c}{\textbf{Normal Error $\downarrow$}} & \multicolumn{3}{c}{\textbf{Normal Accuracy $\uparrow$}}\\ 

\cmidrule(lr){3-5} \cmidrule(lr){6-8} \cmidrule(lr){9-11} \cmidrule(lr) {12-14}
 
 \multicolumn{1}{c}{\textbf{Method}} & \multicolumn{1}{c}{\textbf{\# parameters}} & RMSE & MRE & RMSE log & \textbf{$\delta1 < 1.25$} & \textbf{$\delta2  < 1.25^2$} & \textbf{$\delta3  < 1.25^3$}  & Mean & Median & RMSE & \multicolumn{1}{c}{5.0$^\circ$} & \multicolumn{1}{c}{7.5$^\circ$} & \multicolumn{1}{c}{11.25$^\circ$} \\
%  \multicolumn{1}{c}{\textbf{Method}} & \multicolumn{1}{c}{\textbf{No. of parameters}} & Abs Rel & log10 & RMSE log & \textbf{$\delta1 < 1.25$} & \textbf{$\delta2  < 1.25^2$} & \textbf{$\delta3  < 1.25^3$} \\
 \midrule
 \multicolumn{1}{c}{RectNet \cite{zioulis2018omnidepth}} & \multicolumn{1}{c} {8.9 M}  & \multicolumn{1}{c}{0.646} & \multicolumn{1}{c}{23.786} & \multicolumn{1}{c} {{1.213}} & \multicolumn{1}{c} {{0.247}} &\multicolumn{1}{c} {{0.265}} & \multicolumn{1}{c} {0.283} & \multicolumn{1}{c}{9.84} & \multicolumn{1}{c}{5.49} & \multicolumn{1}{c} {{14.53}} & \multicolumn{1}{c} {{48.84}} &\multicolumn{1}{c} {{56.06}} & \multicolumn{1}{c} {65.85} \\
 
\multicolumn{1}{c}{UResNet \cite{zioulis2018omnidepth}} & \multicolumn{1}{c} {50.8 M}  & \multicolumn{1}{c}{0.097} & \multicolumn{1}{c}{0.487} & \multicolumn{1}{c} {{0.260}} & \multicolumn{1}{c} {{0.424}} &\multicolumn{1}{c} {{0.614}} & \multicolumn{1}{c} {0.768} & \multicolumn{1}{c}{11.50} & \multicolumn{1}{c}{7.18} & \multicolumn{1}{c} {{16.32}} & \multicolumn{1}{c} {{44.50}} &\multicolumn{1}{c} {{49.01}} & \multicolumn{1}{c} {55.73} \\ 

\multicolumn{1}{c}{HoHoNet* \cite{sun2021hohonet}} & \multicolumn{1}{c} {49.5 M} & \multicolumn{1}{c}{0.092} & \multicolumn{1}{c}{0.547} & \multicolumn{1}{c} {{0.228}} & \multicolumn{1}{c} {{0.510}} &\multicolumn{1}{c} {{0.717}} & \multicolumn{1}{c} {0.838} & \multicolumn{1}{c}{X} & \multicolumn{1}{c}{X} & \multicolumn{1}{c}{X} & \multicolumn{1}{c}{X} & \multicolumn{1}{c}{X} & \multicolumn{1}{c}{X} \\ 

 \multicolumn{1}{c}{SliceNet* \cite{Pintore2021SliceNet}} & \multicolumn{1}{c} {79.5 M}  & \multicolumn{1}{c}{0.087} & \multicolumn{1}{c}{0.425} & \multicolumn{1}{c} {{0.232}} & \multicolumn{1}{c} {{0.583}} &\multicolumn{1}{c} {{0.784}} & \multicolumn{1}{c} {0.868} & \multicolumn{1}{c}{X} & \multicolumn{1}{c}{X} & \multicolumn{1}{c}{X} & \multicolumn{1}{c}{X} & \multicolumn{1}{c}{X} & \multicolumn{1}{c}{X} \\

\multicolumn{1}{c}{UBotNet Lite (Ours)} & \multicolumn{1}{c} {38.3 M}  & \multicolumn{1}{c} {0.063} & \multicolumn{1}{c} {{0.403}} & \multicolumn{1}{c} {{0.181}} &\multicolumn{1}{c} {0.657} & \multicolumn{1}{c} { {0.844}} & \multicolumn{1}{c}{0.896} & \multicolumn{1}{c} {8.00} & \multicolumn{1}{c} {{4.19}} & \multicolumn{1}{c} {12.57} &\multicolumn{1}{c} {54.86} & \multicolumn{1}{c} {64.51} & \multicolumn{1}{c}{75.36} \\

\multicolumn{1}{c}{Bifuse* \cite{BiFuse20}} & \multicolumn{1}{c} {212 M} & \multicolumn{1}{c} {0.067} & \multicolumn{1}{c} {{0.345}} & \multicolumn{1}{c} {{0.174}} &\multicolumn{1}{c} {0.646} & \multicolumn{1}{c} { {0.846}} & \multicolumn{1}{c}{0.908} & \multicolumn{1}{c}{X} & \multicolumn{1}{c}{X} & \multicolumn{1}{c}{X} & \multicolumn{1}{c}{X} & \multicolumn{1}{c}{X} & \multicolumn{1}{c}{X}\\

\multicolumn{1}{c}{Panoformer* \cite{shen2022panoformer}} & \multicolumn{1}{c} {20.4 M}  & \multicolumn{1}{c}{0.062} & \multicolumn{1}{c}{0.311} & \multicolumn{1}{c} {0.159} & \multicolumn{1}{c} {0.661} &\multicolumn{1}{c} {0.842} & \multicolumn{1}{c} {0.913} & \multicolumn{1}{c}{X} & \multicolumn{1}{c}{X} & \multicolumn{1}{c}{X} & \multicolumn{1}{c}{X} & \multicolumn{1}{c}{X} & \multicolumn{1}{c}{X} \\

\multicolumn{1}{c}{UNet$_{128}$} & \multicolumn{1}{c} {124 M}  & \multicolumn{1}{c}{\textbf{0.052}} & \multicolumn{1}{c}{\textbf{0.259}} & \multicolumn{1}{c} {0.157} & \multicolumn{1}{c} {{0.641}} &\multicolumn{1}{c} {0.849} & \multicolumn{1}{c} {0.925} & \multicolumn{1}{c}{9.01} & \multicolumn{1}{c}{4.01} & \multicolumn{1}{c} {14.71} & \multicolumn{1}{c} {{54.00}} &\multicolumn{1}{c} {62.58} & \multicolumn{1}{c} {72.68}\\ 

\multicolumn{1}{c}{UBotNet (Ours)} & \multicolumn{1}{c} {133 M}  & \multicolumn{1}{c} {0.054} & \multicolumn{1}{c} {0.271} & \multicolumn{1}{c} {\textbf{0.151}} &\multicolumn{1}{c} {\textbf{0.712}} & \multicolumn{1}{c} {\textbf{0.874}} & \multicolumn{1}{c}{\textbf{0.929}} & \multicolumn{1}{c} {\textbf{7.44}} & \multicolumn{1}{c} {\textbf{3.61}} & \multicolumn{1}{c} {\textbf{12.12}} &\multicolumn{1}{c} {\textbf{56.80}} & \multicolumn{1}{c} {\textbf{67.29}} & \multicolumn{1}{c}{\textbf{78.52}} \\
\bottomrule
% \multicolumn{1}{c}{} &  &  &  &  &  &  &  \\
\end{tabular}%
}
\label{table:benchmarkDepth}
\end{table*}

\begin{figure*}[!ht]
  \centering
  \includegraphics[width=0.92\textwidth]{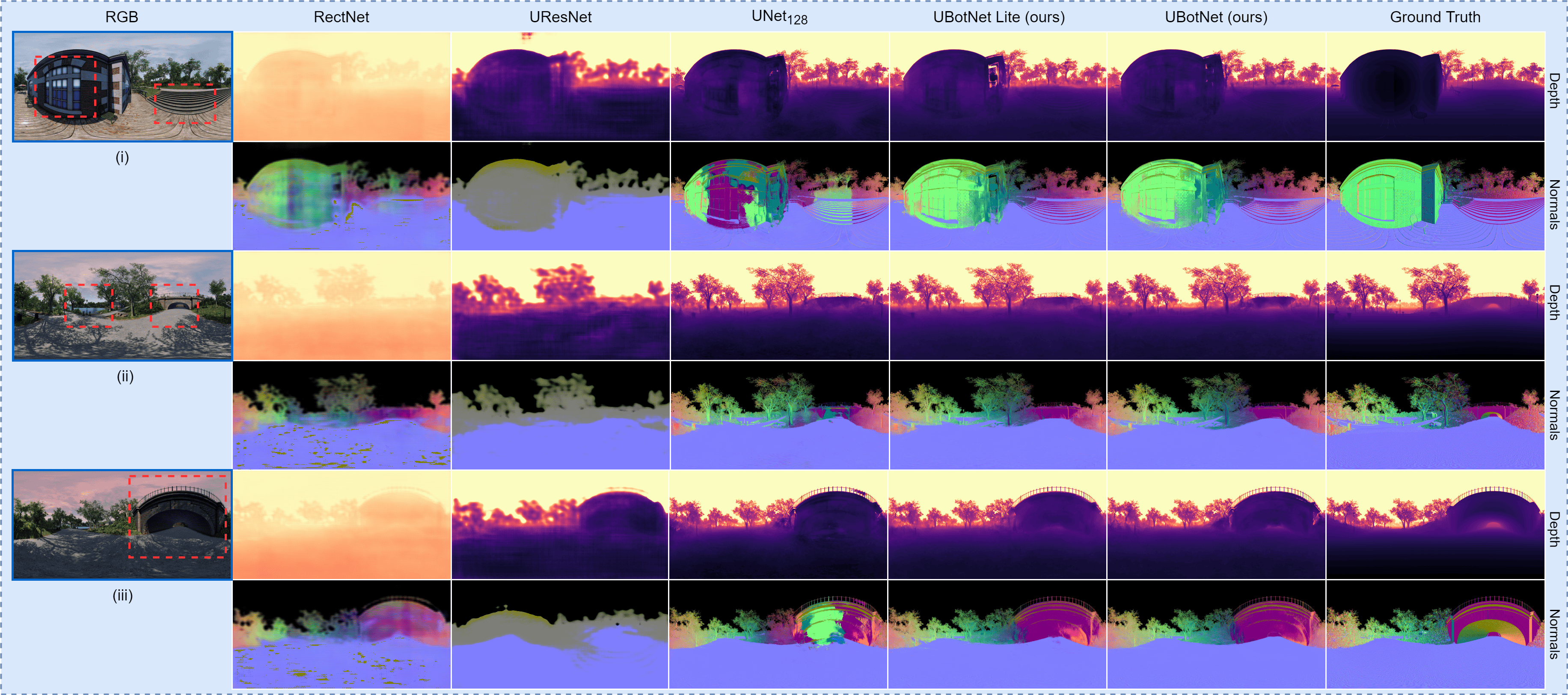}
  \caption{\emph{Qualitative Results from the benchmark on OmniHorizon dataset.} Three different instances of varying depths and lighting conditions are compared between all networks. UBotNet performs consistently better than UNet$_{128}$ and other architectures when estimating depth and normals. UBotNet Lite shows small artefact in depth estimates but still preserves the global context required to learn normals.}
  \label{fig:datasetResults}
  \vspace{-3mm}
\end{figure*}

\subsection{Network Training and Experiments}
\label{section:benchmarks}
We maintain the following setup and configuration for training and testing in all our benchmark, evaluation and ablation experiments.

% The following network training and experiment configurations are maintained for all the benchmark, evaluation and ablation experiments. 
% \paragraph{Training configuration}

\noindent
\textbf{Training configuration.} We used an Nvidia RTX 3090 with 24GB onboard memory for training all network models. The batch size is set to 4 and Adam optimizer \cite{kingma2014adam} is used with a learning rate of $1 \times 10^{-4}$ and decay rate of $1 \times 10^{-5}$. Due to memory constraints, the images were rescaled to a resolution of $512 \times 256$ for training and evaluation. 
% We reserved the training-path 4 (see Figure~\ref{fig:dynamic_lighting}) from City Park scene for test set (1520 images) and used remaining images for training and validation.
% The training-path 4 is isolated from other paths in the scene and contains underpass, stairs, uneven terrain, building and pedestrians which makes it an ideal test set.
%  We used a split of $85:15$ for the training ($19392$) and the validation ($3423$) sets.
All the networks were trained for 40 epochs.
% The networks were trained for 60 epochs with an average time per epoch $\approx$ UResNet($26$ minutes), UBotNet ($ 42 $ minutes), RectNet ($58$ minutes) and UNet$_{128}$ ($58$ minutes).
% \paragraph{Loss Functions}

\noindent
\textbf{Loss Functions.} The networks were    trained to jointly learn both depth and normal information from the input monocular omnidirectional images. We used $\mathcal{L}_{berHu}$ (Reverse-Huber) function \cite{laina2016deeper} as the loss objective for depth and $\mathcal{L}_1$ penalty as the objective function for estimating scene normals .
% \begin{equation}
% % Loss_{sky} = \frac{\sum_{}^{p}(sky_{GT}=sky_{Pred})}{p}    
% \mathcal{L}_{Depth} = \mathcal{L}_{berHu}
% \end{equation}
%
% We used :
% \begin{equation}
% % Loss_{sky} = \frac{\sum_{}^{p}(sky_{GT}=sky_{Pred})}{p}    
% \mathcal{L}_{Normal} = {\mathcal{L}_1}
% \end{equation}
%
The overall loss function for joint learning is, therefore, a sum of both depth and normal objectives:
% \vspace{-2mm}
\begin{equation}
% Loss_{sky} = \frac{\sum_{}^{p}(sky_{GT}=sky_{Pred})}{p}    
\mathcal{L}_{Total} = \mathcal{L}_{Depth} + \mathcal{L}_{Normal}
\end{equation}
% \vspace{-2mm}

\noindent
\textbf{Data Augmentation.}
%Data augmentation is an important tool for generalisation of the networks towards the tasks \cite{DBLP:journals/corr/abs-1712-04621}.
% For generalization of the networks towards the tasks of depth and normal estimation, we used three different augmentation techniques during the training phase.
% Shuffle \cite{eriba2019kornia} and Color Jitter \cite{eriba2019kornia}) of the input images and the third performs geometric transforms (Rotation-based augmentation \cite{pyequilib2021github, jiang2021unifuse}).
% We adapt two techniques augment the color data (Channel Shuffle \cite{eriba2019kornia} and Color Jitter \cite{eriba2019kornia}) of the input images and a third to perform rotation-based augmentation \cite{pyequilib2021github, jiang2021unifuse}.
%
We employ two techniques to augment the color data of the input images, namely Channel Shuffle \cite{eriba2019kornia} and Color Jitter \cite{eriba2019kornia}. Additionally, we use a third technique for rotation-based augmentation \cite{pyequilib2021github, jiang2021unifuse}.

\noindent
\textbf{Baseline Architectures and Evaluation Criteria.}
We evaluated our dataset using various architectures, including SliceNet \cite{Pintore2021SliceNet}, BiFuse \cite{BiFuse20}, Panoformer\cite{shen2022panoformer}, HoHoNet \cite{sun2021hohonet}, UResNet \cite{zioulis2018omnidepth}, RectNet \cite{zioulis2018omnidepth},  UNet$_{128}$ and the proposed UBotNet and UBotNet Lite architecture.
Modifications were applied to the final layers of UResNet and RectNet to enable joint learning of depth and normals. UNet128, employing a base of 128 feature channels extending to 2048 channels, is left similar to the vanilla architecture. Networks like HoHoNet, BiFuse, SliceNet, and Panoformer were left unmodified due to their complexity, training solely for depth estimation. 
Our depth estimation criteria include standard metrics such as  Root Mean Square Error (RMSE), Mean Relative Error (MRE), Root Mean Square Error in log space (RMSE log), and accuracy metrics ($\delta1 , \delta2$ and $\delta3$ with a threshold of $1.25$) \cite{albanis2021, jiang2021unifuse,BiFuse20}.
For normal estimation, evaluation criteria include metrics such as RMSE, mean, median, and accuracy metrics at angles of $5^\circ$, $7.5^\circ$ and $11.25^\circ$ \cite{albanis2021,bae2021estimating,chen2017surface}.

\section{Discussion and Evaluation}
\label{sec:discussion}
\begin{table*}[!ht]
\centering
\caption{\emph{Quantitative results for the ablation study on the OmniHorizon dataset.} Various versions of the dataset are compared by removing the dynamic elements from the scene. VP - Vehicles \& Pedestrians and DL - Dynamic Lighting} 
\resizebox{0.92\textwidth}{!}{%
\begin{tabular}{@{}llllccclllccc@{}} 
\toprule
\multicolumn{1}{c}{} & \multicolumn{3}{c}{\textbf{Depth Error $\downarrow$}} & \multicolumn{3}{c}{\textbf{Depth Accuracy $\uparrow$}} & \multicolumn{3}{c}{\textbf{Normal Error $\downarrow$}} & \multicolumn{3}{c}{\textbf{Normal Accuracy $\uparrow$}}\\ 

\cmidrule(lr){2-4} \cmidrule(lr){5-7} \cmidrule(lr){8-10} \cmidrule(lr) {11-13}
 
 \multicolumn{1}{c}{\textbf{Method}} & RMSE & MRE & RMSE log & \textbf{$\delta1 < 1.25$} & \textbf{$\delta2  < 1.25^2$} & \textbf{$\delta3  < 1.25^3$}  & Mean & Median & RMSE & \multicolumn{1}{c}{5.0$^\circ$} & \multicolumn{1}{c}{7.5$^\circ$} & \multicolumn{1}{c}{11.25$^\circ$} \\
%  \multicolumn{1}{c}{\textbf{Method}} & \multicolumn{1}{c}{\textbf{No. of parameters}} & Abs Rel & log10 & RMSE log & \textbf{$\delta1 < 1.25$} & \textbf{$\delta2  < 1.25^2$} & \textbf{$\delta3  < 1.25^3$} \\
 \midrule

\multicolumn{1}{c}{Static} & \multicolumn{1}{c}{0.055} & \multicolumn{1}{c}{0.293} & \multicolumn{1}{c} {{0.155}} & \multicolumn{1}{c}{0.656} & \multicolumn{1}{c}{0.854} & \multicolumn{1}{c}{0.924} & \multicolumn{1}{c}{7.67} & \multicolumn{1}{c}{3.74} & \multicolumn{1}{c} {12.55} & \multicolumn{1}{c}{56.16} & \multicolumn{1}{c}{66.49} & \multicolumn{1}{c}{77.60} \\ 

\multicolumn{1}{c}{Static + VP}  & \multicolumn{1}{c}{\textbf{0.053}} & \multicolumn{1}{c}{0.289} & \multicolumn{1}{c} {{0.154}} & \multicolumn{1}{c}{\textbf{0.713}} & \multicolumn{1}{c}{0.868} & \multicolumn{1}{c}{0.924} & \multicolumn{1}{c}{7.53} & \multicolumn{1}{c}{3.64} & \multicolumn{1}{c} {{12.26}} & \multicolumn{1}{c}{56.72} & \multicolumn{1}{c}{67.05} & \multicolumn{1}{c}{78.18} \\ 

\multicolumn{1}{c}{Static + VP + DL} & \multicolumn{1}{c}{{0.054}} & \multicolumn{1}{c}{\textbf{0.271}} & \multicolumn{1}{c} {\textbf{0.151}} & \multicolumn{1}{c}{{0.712}} & \multicolumn{1}{c}{\textbf{0.875}} & \multicolumn{1}{c}{\textbf{0.926}} & \multicolumn{1}{c}{\textbf{7.44}} & \multicolumn{1}{c}{\textbf{3.61}} & \multicolumn{1}{c} {\textbf{12.12}} & \multicolumn{1}{c}{\textbf{56.80}} & \multicolumn{1}{c}{\textbf{67.28}} & \multicolumn{1}{c}{\textbf{78.52}}\\

\bottomrule
% \multicolumn{1}{c}{} &  &  &  &  &  &  &  \\
\end{tabular}%
% \footnotesize
% VP - Vehicles + Pedestrians, DL - Dynamic Lighting
}
% \begin{tablenotes}[para,flushleft]
%     VP - Vehicles + Pedestrians, DL - Dynamic Lighting
% \end{tablenotes}

\label{table:ablationStudy}
\end{table*}
\begin{figure*}[!ht]
  \centering
  \includegraphics[width=0.92\textwidth]{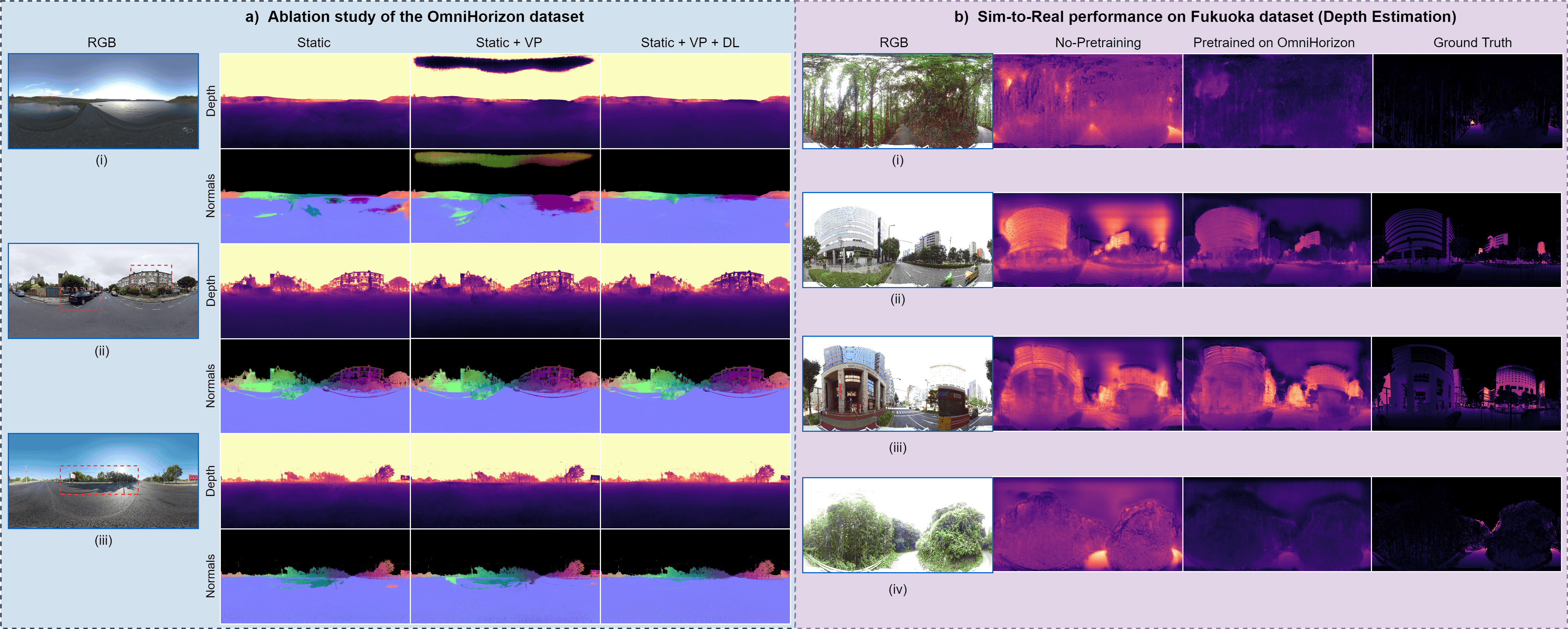}
  \caption{a) \emph{Ablation study of the OmniHorizon dataset.} Comparison for the depth and normal estimation between the various versions of the dataset: Static, Static + VP, and Static + VP + DL. b) \emph{Sim-to-Real performance on Fukuoka dataset.} We compare the performance of depth estimation between the network pre-trained on OmniHorizon and fine-tuned on Fukuoka against the network trained from scratch.}
  \label{fig:ablationResults}
  \vspace{-3mm}
\end{figure*}

\vspace{-1mm}

\subsection{Benchmark Results on OmniHorizon}
% \paragraph{Quantitative results}
\textbf{Quantitative results.}
% Table~\ref{table:benchmarkDepth} shows the quantitative results for the depth and normal estimation for all the networks. The RectNet and UResNet architectures show sub-optimal results on the dataset. RectNet fails to converge after early iterations. On the other hand, other networks demonstrate better outcomes in the benchmark.
% Moreover, UBotNet is consistently better than other architectures including the UNet$_{128}$ on the account of all the metrics except for MRE. For normal metrics, we observe a performance uplift of $14.92\%$ for normal error and $4.45\%$ for normal accuracy between UNet$_{128}$ and UBotNet.
% We observe that UBotNet Lite (38.3 M) performs slightly lower than UNet$_{128}$ (124 M) for certain metrics but shows better results for normal metrics while having 70\% less parameters when compared to the UNet$_{128}$.
%
Quantitative results for depth and normal estimation across all networks are shown in Table~\ref{table:benchmarkDepth}. RectNet and UResNet architectures exhibit suboptimal performance, with RectNet failing to converge after early iterations. Conversely, other networks, including UBotNet, show superior outcomes in the benchmark. Notably, UBotNet consistently outperforms other architectures, including UNet$_{128}$, across all metrics except for Mean Relative Error (MRE). In terms of normal metrics, UBotNet demonstrates a performance improvement of 14.92\% for normal error and 4.45\% for normal accuracy compared to UNet$_{128}$. UBotNet Lite (38.3M params) performs slightly lower than UNet$_{128}$ (124M params) for specific metrics but shows better results for normal metrics while having 70\% fewer parameters.
% than UNet128.

% \paragraph{Qualitative results}

% Table~\ref{table:benchmarkDepth} shows the quantitative results for the depth and normal estimation for all the networks. The RectNet and UResNet architectures show sub-optimal results on the dataset. RectNet fails to converge after early iterations. On the other hand, both UBotNet and UNet$_{128}$ demonstrate better outcomes in the benchmark. It is evident from the results that both architectures greatly benefit from the skip-connections.
% Moreover, UBotNet is consistently better than other architectures including the UNet$_{128}$ on the account of all the metrics except for MRE. For normal metrics, we observe a performance uplift of $14.92\%$ for normal error and $4.45\%$ for normal accuracy between UNet$_{128}$ and UBotNet.
% We observe that UBotNet Lite (38.3 M) performs slightly lower than UNet$_{128}$ (124 M) for certain metrics but shows better results for normal metrics while having 70\% less parameters when compared to the UNet$_{128}$.

\textbf{Qualitative results.} 
In Figure~\ref{fig:datasetResults}, visual comparisons for depth\footnote{Depth maps have been normalised for visualisation purposes.} and normals across various architectures are presented, showcasing their validation against Ground Truth (GT) data. The first image highlights structures in close proximity, where UBotNet exhibits superior depth estimation for building windows and stairs compared to UNet$_{128}$. Additionally, UBotNet provides more accurate normal estimates for the stairs and building structure. The second and third images focus on distant elements, testing the networks' ability to identify trees and underpass structures in shadows. Note that UNet$_{128}$ struggles to identify the distant part of the tunnel in the third image, while UBotNet successfully detects it. Furthermore, UBotNet and UBotNet Lite shows estimates closer to GT and outperforms UNet$_{128}$, which completely fails, in estimating normals for the underpass structure in both images. This validates that the proposed architectures demonstrate strong performance in normal estimation, benefiting from the global context extracted by the Multi-Head Self-Attention (MHSA) from the encoder features. Qualitative results for other networks (marked with * in Table \ref{table:benchmarkDepth}) are available in the supplementary material.

\begin{figure*}[!ht]
  \centering
  \includegraphics[width=0.92\textwidth]{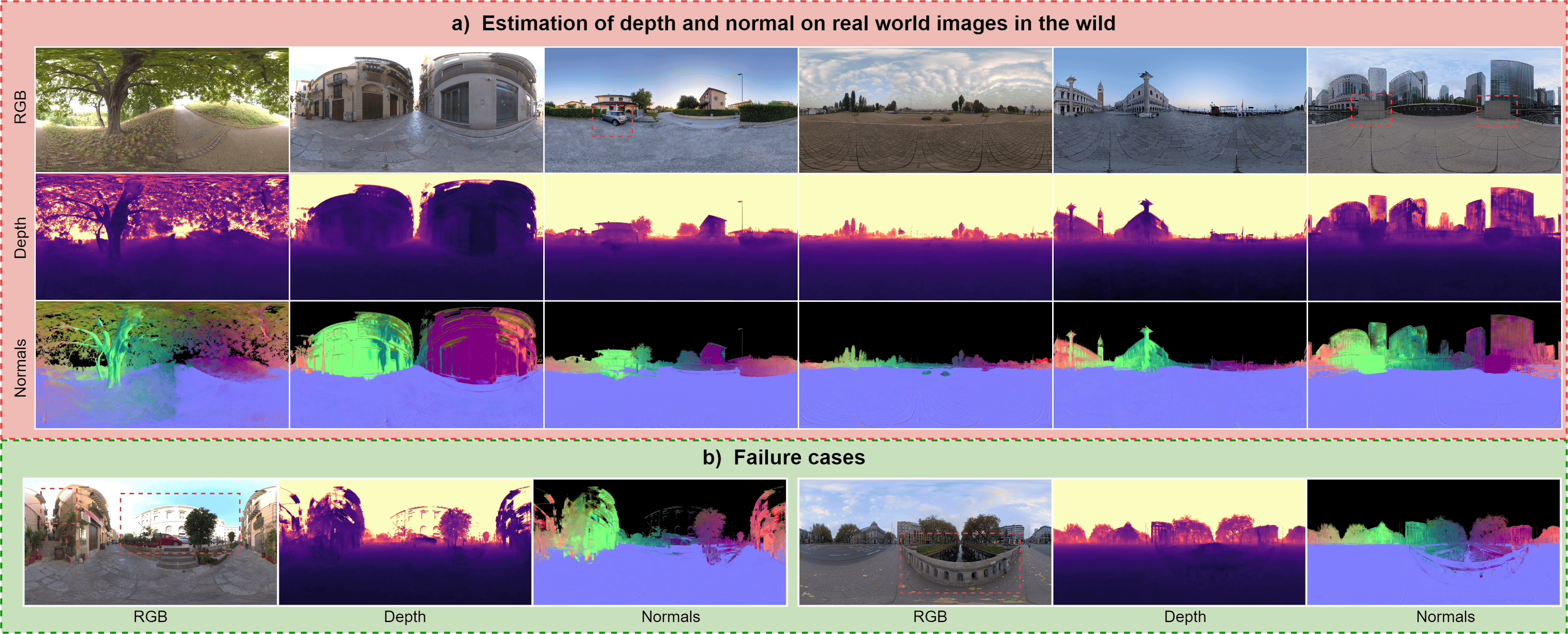}
  \vspace{-2mm}
  \caption{a) \emph{Predictions on the real-world images in the wild}. Depth and normals estimated from real-world images representing the diverse outdoor scenarios. b) \emph{Failure cases.} Network fails to estimate depth and normals in scenarios with overexposed regions. It also fails to recognize vertical upright structure such as the bridge railing.}
  % \caption{a) \emph{Predictions on the real-world images in the wild}. Depth and normals estimated from real-world images representing the diverse outdoor scenarios. The images show different sky conditions including overexposed sky, cloudy and clear sky with no clouds. It also features man-made constructions, vehicles and vegetation. b) \emph{Failure cases.} Network fails to estimate depth and normals in scenarios with overexposed regions. It also fails to recognize vertical upright structure such as the bridge railing.}
  \label{fig:realworldResults}
  \vspace{-4mm}
\end{figure*}

\subsection{Ablation Study}
% We perform the ablation study to address an important question : \textit{Does context matter in outdoor scenarios?}
% It evaluates the contribution of our dataset in terms of the dynamic components: vehicles and pedestrians (VP), and dynamic lighting (DL). Therefore, we create two additional versions of the dataset. First, we generate the static version of the dataset which includes only static meshes and lacks any dynamic components. The second version of the dataset includes pedestrians and vehicles but lacks dynamic lighting component. Note that the full OmniHorizon dataset consists of all dynamic components. 
Our ablation study aims to address a key question: \emph{Does context matter in outdoor scenarios?} It assesses the impact of dynamic components, specifically vehicles and pedestrians (VP), and dynamic lighting (DL), on the dataset. We create two additional dataset versions for comparison: a static version with only static meshes and no dynamic components, and a version with pedestrians and vehicles but without dynamic lighting. Note that the full OmniHorizon dataset includes all dynamic components. 
Table~\ref{table:ablationStudy} shows the comparison between the various versions of the dataset using the UBotNet architecture. We observed incremental gain in the performance with the addition of dynamic components, specifically for depth accuracy and normal metrics. Figure~\ref{fig:ablationResults} shows the visual results from the ablation study. For the first image, Static version struggle with the normal estimation for water surface while the Static+VP version faces issues with lighting and normal estimation. Differences in the vehicle on the left and building are observed in the second image between the static version and others. 
% The inclusion of vehicles to the dataset reflects in better results in the real-world examples.
% Finally, the last image demonstrates the shadow artifacts present in normal maps in both Static and Static+VP version which are absent in the result from the full dataset.
The last image highlights shadow artifacts present in normal maps for the static and static+VP versions, absent in the results from the full dataset.
The above visual differences between the different versions of the dataset demonstrate the importance of context and dynamic elements in the outdoor scenarios. Moreover, the absence of these elements could impair the capacity of neural networks to make accurate predictions.

\begin{table}[!ht]
\centering
\caption{\emph{Quantitative results for Sim-to-Real performance on Fukuoka dataset after pre-training on OmniHorizon.}}
\resizebox{0.4\textwidth}{!}{%
\begin{tabular}{@{}lllllll@{}}
\toprule
\multicolumn{1}{c}{} & \multicolumn{3}{c}{\textbf{Depth Error $\downarrow$}} & \multicolumn{3}{c}{\textbf{Depth Accuracy $\uparrow$}}\\

\cmidrule(lr){2-4} \cmidrule(lr){5-7}
 
 \multicolumn{1}{c}{\textbf{Model}} & RMSE & MRE & RMSE log  & \multicolumn{1}{c} {$\delta1$} & \multicolumn{1}{c}{$\delta2$} & \multicolumn{1}{c}{$\delta3$} \\
%  \multicolumn{1}{c}{\textbf{Method}} & \multicolumn{1}{c}{\textbf{No. of parameters}} & Abs Rel & log10 & RMSE log & \textbf{$\delta1 < 1.25$} & \textbf{$\delta2  < 1.25^2$} & \textbf{$\delta3  < 1.25^3$} \\
 \midrule
\multicolumn{1}{c}{UBotNet*}  & \multicolumn{1}{c}{0.036} & \multicolumn{1}{c}{0.633} & \multicolumn{1}{c} {{0.307}} & \multicolumn{1}{c}{0.265} & \multicolumn{1}{c}{0.497} & \multicolumn{1}{c} {{0.664}} \\ 
 \midrule
\multicolumn{1}{c}{UNet}  & \multicolumn{1}{c}{0.036} & \multicolumn{1}{c}{0.615} & \multicolumn{1}{c} {0.301} & \multicolumn{1}{c} {0.422} &\multicolumn{1}{c} {0.647} & \multicolumn{1}{c} {0.771} \\

    \multicolumn{1}{c}{Bifuse}  & \multicolumn{1}{c}{0.034} & \multicolumn{1}{c}{0.638} & \multicolumn{1}{c} {0.289} & \multicolumn{1}{c} {\textbf{0.447}} &\multicolumn{1}{c} {\textbf{0.659}} & \multicolumn{1}{c} {0.773} \\ 

\multicolumn{1}{c}{UBotNet} & \multicolumn{1}{c}{\textbf{0.029}} & \multicolumn{1}{c}{\textbf{0.611}} & \multicolumn{1}{c} {\textbf{0.271}} & \multicolumn{1}{c} {{0.424}} &\multicolumn{1}{c} {{0.655}} & \multicolumn{1}{c} {\textbf{0.782}} \\

\bottomrule
% \rule{0pt}{4ex}
\multicolumn{7}{l}{*\footnotesize{trained only on Fukuoka dataset}}\\

% \multicolumn{1}{c}{} &  &  &  &  &  &  &  \\
\end{tabular}%
}
\vspace{-4mm}
\label{table:sim2real}
\end{table}

\subsection{Sim-to-Real Transfer}
We evaluate the simulation-to-real domain transfer performance of our method on a real-world dataset - Fukuoka\cite{mozos2019}. To achieve this task, we pretrain the UBotNet on our dataset and fine-tune it on Fukuoka for the task of depth estimation. Note that Fukuoka dataset does not provide ground truth for normal data and therefore we only evaluate the depth estimates. Additionally, Fukuoka only provides 650 images for training compared to the OmniHorizon with 24,335 samples. Hence, we pre-train the network only on 2K samples ($<$ 10\% of our dataset). Table~\ref{table:sim2real} summarizes the performance comparison between the networks pre-trained on our dataset and that trained on Fukuoka from scratch. We noted better performance of the pretrained network specifically for depth accuracy, where we see a gain of $12.2\%$. We also observed more accurate depth maps estimated from the test images when compared to training from scratch on Fukuoka as shown in Figure~\ref{fig:ablationResults}. When trained from scratch, the network struggles notably with vegetation. On the other hand, it benefits from a better understanding of scenes with complex vegetation when it was pre-trained on OmniHorizon. 
Interestingly, we also observe a similar trend for other networks that were trained first on OmniHorizon.

% \vspace{-3mm}
\subsection{Testing on the Real-world Images In-the-wild}
The real-world omnidirectional images have been curated from the Polyhaven website \cite{polyhaven2022} for testing the trained network on the images in the wild. We selected images that represent diverse outdoor scenarios cluttered with various objects and captured during different time of day settings. Figure~\ref{fig:realworldResults} shows depth and normals estimated by UBotNet from the images. The images illustrate the ability of the network to estimate depth at a large range in various settings. Our network learns high level details from the vegetation (images 1, 3 and 4). This is reflected in the image 1 where the network was able to recognize the large tree in the foreground along with the walking path. It also captures the details from the cars in image 3. The network was able to identify sky region in cases with full clouds (image 2) and clean sky with no clouds(image 3 and 5). This demonstrates the advantage of the including various cloud formations and time of day settings in the dataset. The final image which shows a skating area is a good example of the ability of UBotNet to estimate normals of two upright structures (highlighted in \textcolor{red}{red}) in front of the buildings with a texture similar to the concrete floor. It highlights the capacity of the network to learn information in a global context to understand the orientation of normal surfaces. Overall, the network demonstrates promising results for the estimation of depth and normal on real-world images. We show additional results in the supplementary material.

\subsection{Limitations}
There are specific scenarios where sunlight may overexpose parts of a scene while underexposing others. In such instances, the network struggles to correctly estimate depth and normals for the overexposed parts of the scene. Additionally, the network occasionally misinterprets vertical elements like handrails and bridge supports. Figure~\ref{fig:realworldResults} shows both such challenging scenarios where our method compromised. We discuss the assumptions of our dataset in supplementary material.
\vspace{-2mm}
\section{Conclusion}
We presented a new dataset called OmniHorizon and a hybrid architecture called UBotNet for depth and normal estimation in diverse outdoor scenarios. 
Firstly, our dataset includes diverse outdoor spaces and also dynamic scene participants such as pedestrians and vehicles.
% We noticed that dynamic components in the dataset are critical and not including them results in sub-optimal inferences on in-the-wild real-world examples.
% Our dataset includes pedestrians and vehicles as they are critical dynamic components which, if not included, may result in suboptimal results when dealing with real-world examples. 
% We also discussed the challenges involved in representing outdoor spaces such as buildings, vegetation, sky and varying lighting conditions and approaches to overcome those.
Secondly, our UBotNet, based on U-Net and Bottleneck transformer, trained on the OmniHorizon dataset demonstrated significantly improved and scene-consistent normal estimation against the vanilla U-Net architecture. Furthermore, we presented UBotNet Lite, a smaller version of the network that retains respectable depth and normal accuracy while having only 30\% of the network parameters.
We outlined the benefits of pre-training network on OmniHorizon and fine-tuning it on Fukuoka dataset.
Finally, we demonstrated the application of our model trained on OmniHorizon for estimating the depth and normals of real-world outdoor omnidirectional images in-the-wild.

% We are very excited to release our dataset and models to the research community. 

% We presented a new dataset in this work called OmniHorizon for depth and normal estimation in outdoors scenarios. This work addressess different challenges and aspects of representing outdoor spaces which include buildings, vegetation, sky and varying lighting conditions. Pedestrians and vehicles are important dynamic components, if not included may result in suboptimal results while dealing with real-world examples. We also proposed a new architecture UBotNet based on UNet and Bottleneck transformer for scene-consistent normals. Finally, we use the trained networks to estimate depth and normals from real-world omnidirectional images in the wild. The results are promising and show the potential application of our dataset for zero-shot learning.
% Finally, we have demonstrated the application of training networks on our dataset by estimating depth and normals from the real-world omnidirectional images in the wild.
% We are very excited to release our dataset and network models, and are confident that researchers will find it useful for tackling some of the challenging scene understanding problems.

% \newpage
%%%%%%%%% REFERENCES

\newpage

% \begin{figure*} [H]
%     \captionsetup{type=figure}
%     \includegraphics[width=\textwidth]{Figures/SupplementaryIntro.png}
%     \captionof{figure}{\emph{Example of RGB, depth and normals for all the scenes from the OmniHorizon dataset.}}
%     \label{fig:exa}
% \end{figure*}
% We further discuss the considerations and certain assumptions that we made in order to render the dataset. Finally, we demonstrate additional results for depth and normal estimation from real-world images in the wild.
%-------------------------------------------------------------------------
\setcounter{section}{0}
\section{Supplementary Material}
In this supplementary material, we discuss our approach on generating the OmniHorizon dataset in Unreal Engine 4. We elaborate on the factors and certain assumptions that we made in order to render the dataset. Additionally, we discuss about training the UBotNet on indoor datasets and architecture choices. Finally, we demonstrate additional results for depth and normal estimation from real-world images in the wild.

\subsection{Depth clamping}
\begin{figure*} [!h]
    \centering
    \includegraphics[width=\textwidth]{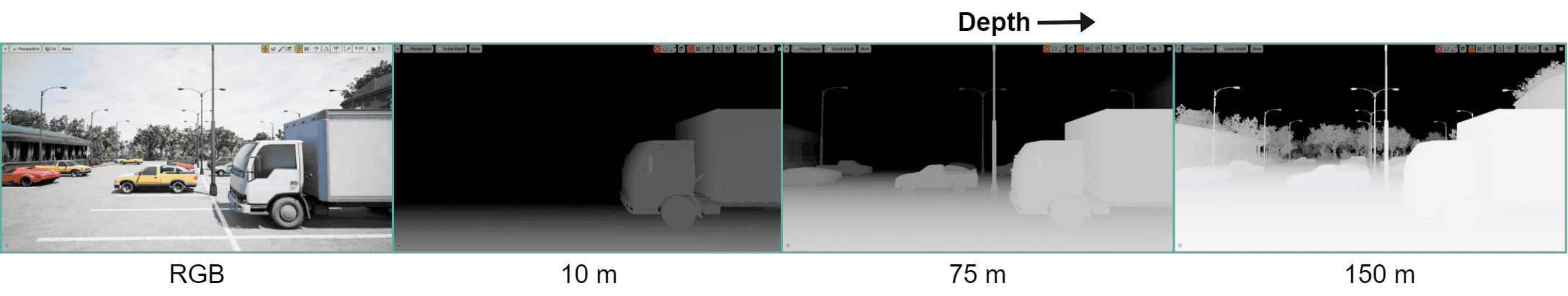}
    \caption{\emph{Depth clamping experiment.} Comparison between various depth ranges after clamping to a specific range: 10 m, 75 m and 150 m. Inverted depth maps are shown for better visualization.}
    \label{fig:depth_clamping}
\end{figure*}

Rendering engines such as Unreal Engine 4 work with a larger depth range compared to that captured by physical sensors. However, we were interested in exploring the range of depth information that can be used for covering a wide range of objects in outdoor scenarios. This motivated us to simulate the limitations of the physical sensors and restrict the depth range to 150 m, similar to the Fukuoka dataset \cite{mozos2019}.
The engine places the far plane at infinity, which results in depth values being generated for extremely distant objects. To avoid this, we modify the depth material to visualise the impact of constraining the depth to a maximum specified value. We show the results for the clamping of depth at a range of 10m, 75m and 150m in Figure~\ref{fig:depth_clamping}. At a depth of 10 m, only the truck is visible. When the depth range is raised to 75 m, cars and building start to appear in the background. At 150 m, the trees and most of the background are visible.
% Finally, at 500 m, the depth map begins to get saturated as the majority of observable objects in the scene are now represented. There are no major changes in depth above 500 meters.
By limiting the depth in outdoor environments, it is possible to focus solely on nearby items, or, depending on the application, on distant objects as well. 

\subsection{View-space vs world-space normals}
The view space normals are calculated relative to the camera orientation, whereas the world space normals are calculated with respect to the global axes of the scene. The normals in view space are desired when using a perspective camera as they are tied to the camera pose (extrinsic parameters). However, the panoramic image is obtained by rotating the camera on both the horizontal and vertical axis in increments of fixed angle steps (5°), followed by merging the multiple views.

\begin{figure}[!h]
    \centering
    \begin{subfigure}{0.235\textwidth}
    \includegraphics[width=\textwidth]{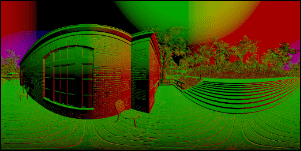}
    \caption{view-space normals}
    \label{fig:viewspace}
    \end{subfigure}
    % \hfill
    \begin{subfigure}{0.235\textwidth}
    \includegraphics[width=\textwidth]{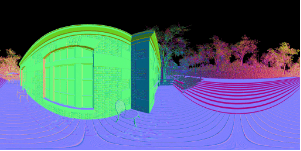}
    \caption{world-space normals}
    \label{fig:worldspace}
    \end{subfigure}
    \vspace{-7mm}
    \caption{\emph{Comparison between view-space and world-space normals.} The normals captured in view-space appear as gradient with lack of clear distinction between the basis vectors. Normal maps recorded in world-space follow a consistent coordinate system.}
    \label{fig:viewVSworld}
\end{figure}
\begin{figure}[!h]
    \centering
    \includegraphics[width=0.45\textwidth]{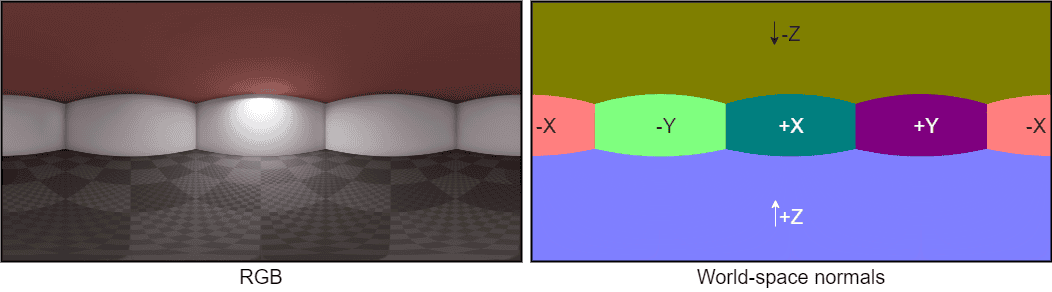}
    % \vspace{-3mm}
    % \includegraphics[width=0.5\textwidth]{Figures/view-space normal.png}
    \caption{\emph{Convention for the world-space normals.}}
    \label{fig:normalsConvention}
\end{figure}

\begin{figure*} [!h]
    \centering
    \includegraphics[width=\textwidth]{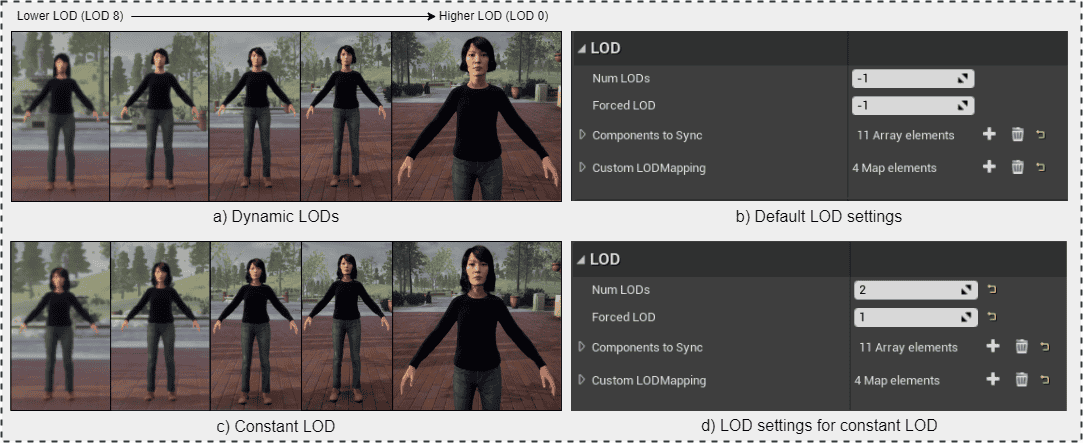}
    \caption{\emph{Dynamic LODs vs Constant LOD.} a) The Dynamic LOD system loads different meshes with various level of details based on the proximity to camera. This however results in sudden poping up of the meshes which generates artefacts in the data. b) Default LOD settings used by the engine. c) The modified LOD system is used to maintain LODs at a fixed LOD so that the avatar's appearance is unaffected by distance. d) The LOD of the character is locked to 1 using Forced LOD.}
    \label{fig:lod}
\end{figure*}

Since the coordinate system is relative to the camera in view space, it also gets modified with the rotation. This results in a gradient of normals with no basis vectors. The normals obtained in world space are absolute and independent of camera pose. Figure~\ref{fig:viewVSworld} shows the difference between the view-space and world-space normals. Therefore, we captured the normals in world space as it was consistent for both within and between the scenes. We show the convention used for the world-space normals in Figure~\ref{fig:normalsConvention}.

\begin{figure}[!h]
    \centering
    % \begin{subfigure}{0.235\textwidth}
    \includegraphics[width=0.46\textwidth]{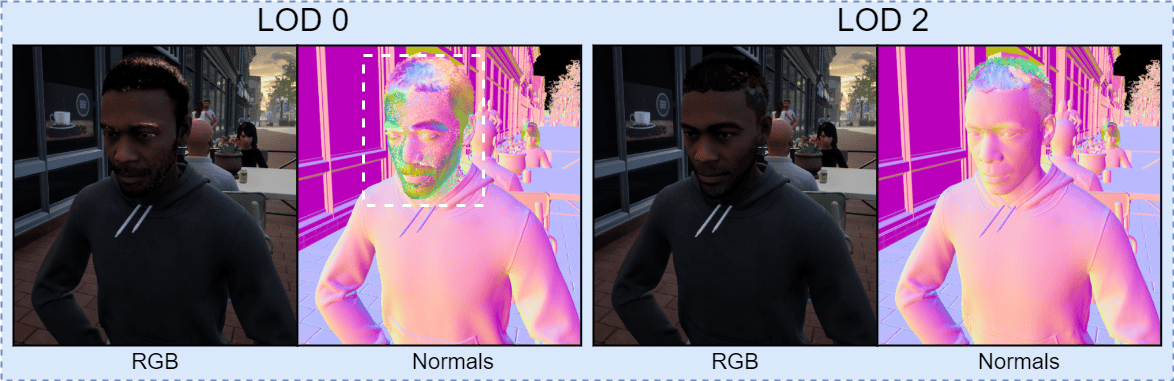}
    \caption{\emph{Artefacts in normal maps for facial hairs.} When the camera is very close to the characters, the engine uses additional detail meshes for characters with facial hair at the highest LOD level (LOD 0). As a result, artefacts appear in the normal maps.We use LOD 1 or 2 for such characters.}
    \label{fig:normal_issues_metahumans}
    % \vspace{-5mm}
    % \end{subfigure}
\end{figure}

\subsection{Virtual Avatars}
As discussed in main paper, we utilised Metahumans \cite{metahumans2022} for the virtual avatars in the scene. We have used premade MetaHumans available in the Quixel bridge. It allowed us to bring in highly detailed characters and more diversity in the pedestrians. But there were certain challenges while using the Metahumans for the dataset. They are generated with multiple level of details (LODs) for perfomance optimisation. As a result, there would be sudden popups and other artifacts when the camera is approaching a character. Figure~\ref{fig:lod} illustrates how the character hair and details change when the camera is approaching the character. Lower LOD level (LOD 8) indicates lowest detailed polygon mesh with no advanced features such as detail normal maps or hairs. The higher LOD level (Level 0/1) has higher polygons with extra detail maps for the skin and hair grooming system. Additionally, we also observed artifacts in the normal maps for the characters with detailed grooming such as facial hair. Figure~\ref{fig:normal_issues_metahumans} shows the issues with the normal maps of a character in the region with facial hair. For such characters we used LOD 1 or LOD 2 to resolve the problems.

\begin{figure*} [!h]
    \centering
    \includegraphics[width=\textwidth]{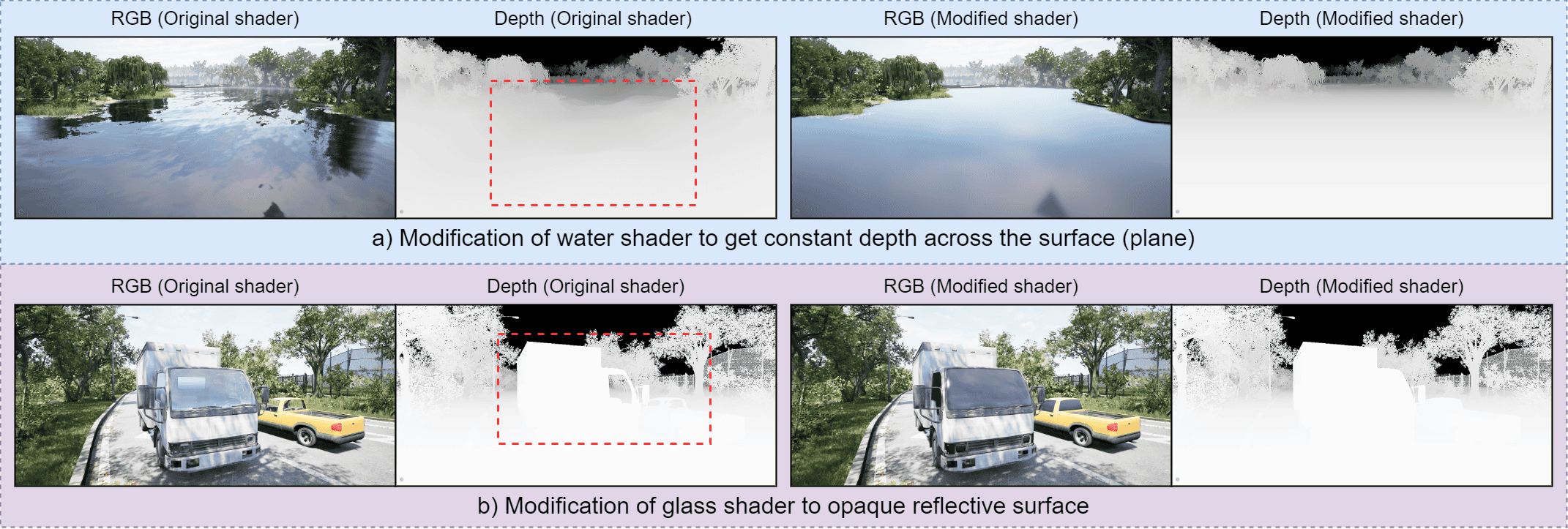}
    \caption{\emph{Assumptions for the dataset.} a) Modification of water shader to achieve constant depth across the surface of the water. b) Modification of glass shader into opaque reflective surface which hides the interior parts of the vehicles.}
    \label{fig:assumptionsForTheDataset}
\end{figure*}

\begin{figure*} [!h]
    \centering
    \includegraphics[width=0.75\textwidth]{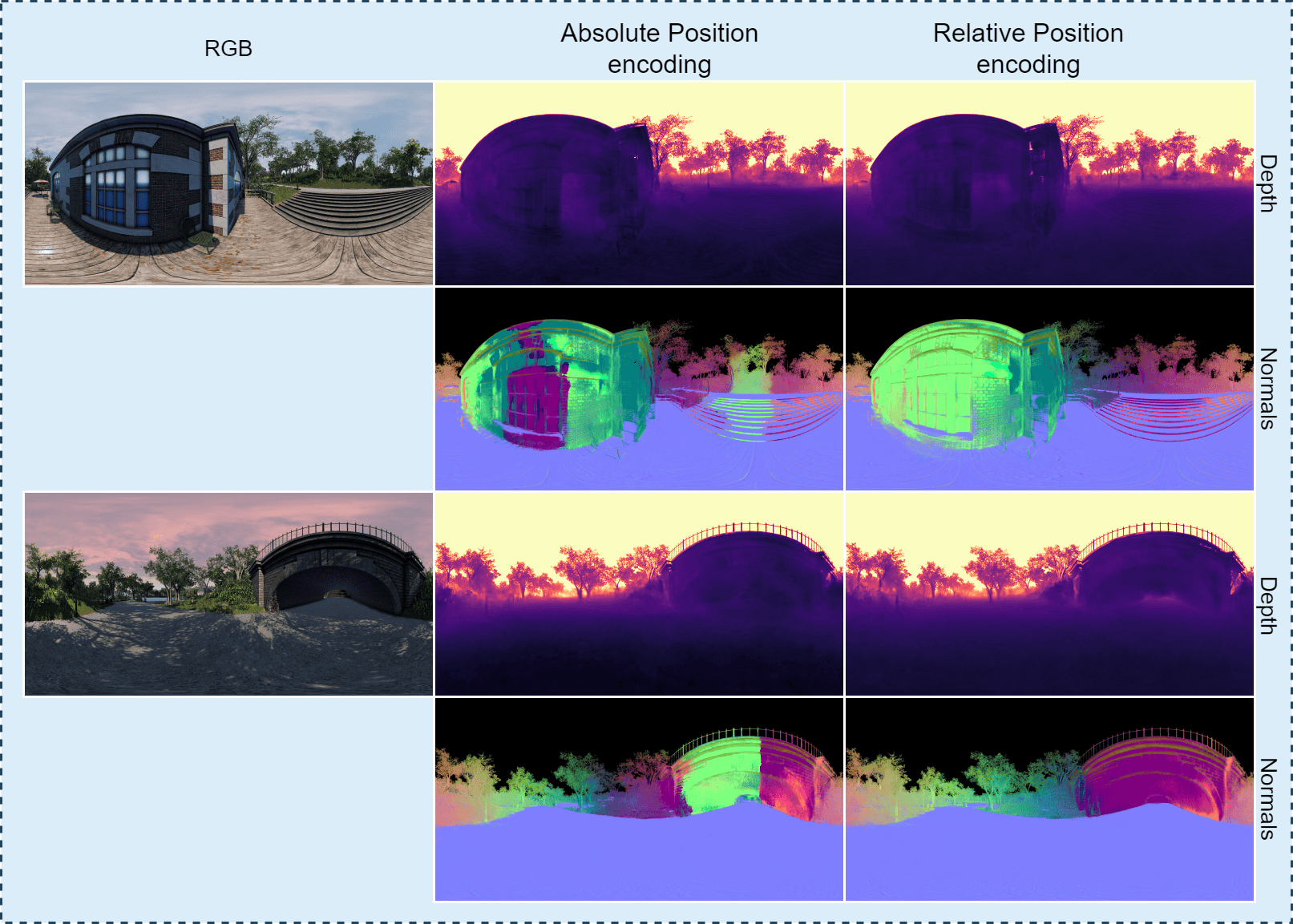}
    \caption{\emph{Comparison between Abs. and Rel. positional embedding.} Absolute positional embedding loses the context required for learning the normals when used for self-attention.}
    \label{fig:absPosEmbedding}
\end{figure*}

\subsection{Assumptions in the Dataset}
Our dataset renders several realistic outdoor and indoor environments with dynamic scene components. While curating this dataset, we made certain assumptions especially about the outdoor scenes which we list below:
\begin{enumerate}
\item
The sky is assumed to be situated at infinitely large distance from the camera, and is represented as a spherical mesh of large radius encompassing the entire scene. Additionally, normals are not rendered for the sky region. It is represent using black which indicates invalid normal values. This allows us to distinguish sky from other regions in the scene.
\item
Transparent and transluscent materials such as water, windows of the buildings and windshields of vehicles are replaced with fully reflective materials. We observed that inferring depth of such materials from color images is challenging and this
%Transparent materials use scene depth with complex shaders that cannot be easily inferred from monocular scene color image. Note that this 
limitation, for example, also applies to real-world datasets captured using lidars \cite{diode_dataset}. Figure~\ref{fig:assumptionsForTheDataset} depicts the limitation of using transparent and translucent materials in the dataset. The original water shader in the scene was designed in such a way that it acted as a see-through material in case of depth. As a result, the depth map captures the terrain hidden underneath the water surface. We modified the the water shader to a reflective surface and thus depth is correctly rendered as a planar surface. We observed a similar case for the glass shader used for windows in the vehicles. The vehicles indeed have detailed indoors but due to reflections on the glass, the inside is not clearly visible. However, the depth map has much cleaner view of the indoors. To avoid this conflict of information, we use fully opaque and reflective materials for the windows.
%We have avoided using transparent and translucent materials in the dataset since such materials use scene depth with complex shaders which cannot be inferred from the scene color image. We replaced them with fully reflective materials. Example of this include the water in the lake of CityPark scene, windows of the buildings and window shields of vehicles.
%It is interesting to note that such limitations also apply to real-world datasets captured using LIDAR sensors \cite{diode_dataset}. 
\end{enumerate}

\section{UBotNet}
\paragraph{UBotNet for Indoor datasets.} In the main paper, we discussed about the UBotNet architecture and the results from training on the OmniHorizon dataset. We additonally trained UBotNet on real-world indoor dataset Pano3D\cite{albanis2021} to validate the performance of the network on other datasets. Pano3D is proposed as a modification of Matteport3D\cite{Matterport3D} and Gibson3D\cite{Xia_2018_CVPR}. We used the official splits provided by the authors for Matterport3D for training and validation. For, Gibson, we used the \emph{GibsonV2 Full Low Resolution} for training and validated on Matterport. All the images used for training were of 512 x 256 resolution. We used the loss function and training parameters outlined in our main paper. We trained UBotNet Lite on the both the datasets for 60 epochs.
% In the main paper, we discussed about the UBotNet architecture and the results from training on the OmniHorizon dataset. We additonally trained UBotNet on real-world indoor dataset Pano3D\cite{albanis2021} to validate the performance of the network on other datasets. We used the official splits provided by the authors for Matterport3D\cite{Matterport3D}. All the images used for training were of 512 x 256 resolution. We used the loss function and training parameters outlined in our main paper. The networks were trained for 60 epochs.
\begin{table}[!ht]
\centering
\caption{\emph{Quantitative results for depth estimation using UBotNet Lite validated on indoor dataset - Matterport3D.}}
\resizebox{0.45 \textwidth}{!}{%
\begin{tabular}{@{}lllllll@{}}
\toprule
\multicolumn{1}{c}{} & \multicolumn{3}{c}{\textbf{Depth Error $\downarrow$}} & \multicolumn{3}{c}{\textbf{Depth Accuracy $\uparrow$}}\\

\cmidrule(lr){2-4} \cmidrule(lr){5-7}
 
 \multicolumn{1}{c}{\textbf{Dataset}} & RMSE & MRE & RMSE log  & \multicolumn{1}{c} {$\delta1$} & \multicolumn{1}{c}{$\delta2$} & \multicolumn{1}{c}{$\delta3$} \\
%  \multicolumn{1}{c}{\textbf{Method}} & \multicolumn{1}{c}{\textbf{No. of parameters}} & Abs Rel & log10 & RMSE log & \textbf{$\delta1 < 1.25$} & \textbf{$\delta2  < 1.25^2$} & \textbf{$\delta3  < 1.25^3$} \\
 \midrule
\multicolumn{1}{c}{Matterport3D}  & \multicolumn{1}{c}{0.639} & \multicolumn{1}{c}{{0.142}} & \multicolumn{1}{c} {{0.064}} & \multicolumn{1}{c}{0.817} & \multicolumn{1}{c}{0.952} & \multicolumn{1}{c} {{0.981}} \\ 

\multicolumn{1}{c}{Gibson 3D}  & \multicolumn{1}{c}{{0.591}} & \multicolumn{1}{c}{0.154} & \multicolumn{1}{c} {{0.061}}  & \multicolumn{1}{c}{{0.830}} & \multicolumn{1}{c}{{0.965}} & \multicolumn{1}{c} {{0.986}}\\ 

\bottomrule
% \multicolumn{1}{c}{} &  &  &  &  &  &  &  \\
\end{tabular}%
}
\label{table:matterport3d}
\end{table}

% 'a1': 0.8308489918708801,
%  'a1_angle_error': 27.407848358154297,
%  'a2': 0.9656582474708557,
%  'a2_angle_error': 28.21405601501465,
%  'a3': 0.9860982894897461,
%  'a3_angle_error': 28.857221603393555,
%  'a4_angle_error': 29.758447647094727,
%  'a5_angle_error': 30.1363525390625,
%  'abs': 0.3276004195213318,
%  'abs_rel': 0.1545877307653427,
%  'log10': 0.0425298735499382,
%  'mean_angle_error': 74.38201141357422,
%  'median_angle_error': 89.9997329711914,
%  'rmse': 0.591411292552948,
%  'rmse_angle_error': 92.66288757324219,
%  'rmsle': 0.06162891164422035,
%  'sq_rel': 0.12692393362522125

\begin{figure*} [!h]
    \centering
    \includegraphics[width=0.85\textwidth]{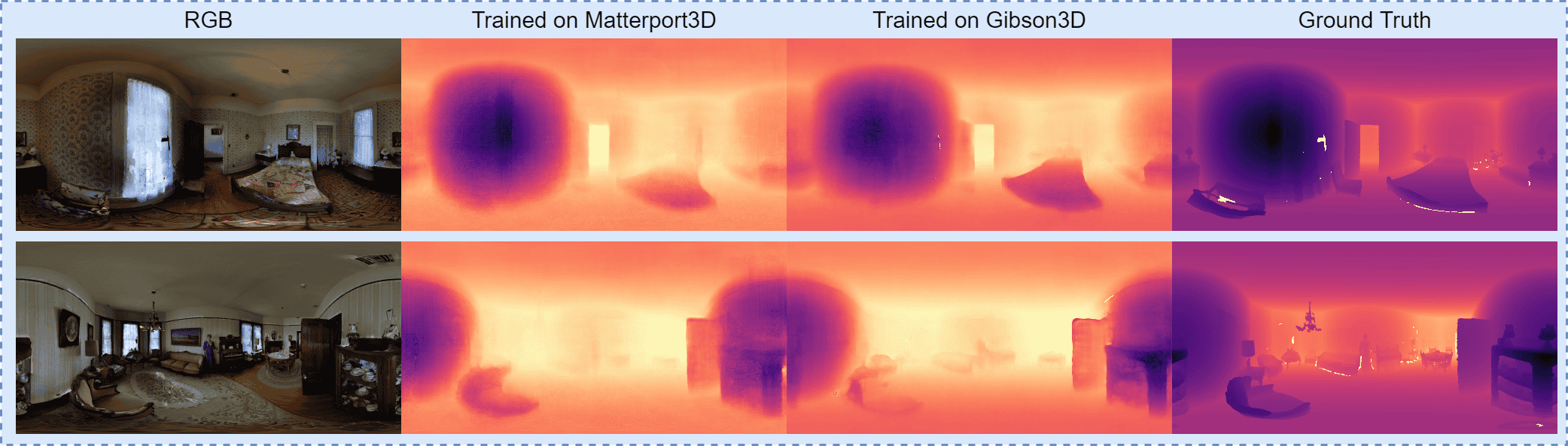}
    \caption{\emph{Qualitative results for UBotNet Lite trained on Indoor datasets - Matterport3D and Gibson3D.}}
    \label{fig:matterportResults}
\end{figure*}

Table~\ref{table:matterport3d} shows the quantitative results for the task of depth estimation by UBotNet Lite evaluated on Matterport3D. We also show the qualitative results for the validation task in Figure~\ref{fig:matterportResults}. We observed better performance in overall metrics and the visual results when the network is trained on the Gibson3D.

\paragraph{Absolute vs Relative positional encoding.} We utilised relative positional encoding  \cite{Srinivas_2021_CVPR} for self-attention in our proposed UBotNet architecture. We compare it against the absolute positional embeddings and show the quantitative results in Table~\ref{table:positionalEmbedding}. The absolute positional embeddings perform inferior to the relative positional embeddings used for self-attention. Moreover, the differences are more prominent in case of normal estimation. This is reaffirmed by the visual differences shown in Figure~\ref{fig:absPosEmbedding}. The network loses the context required for learning the consistent representation of the normals. It behaves similar to the UNet$_{128}$ network discussed in the main paper.

\paragraph{Network architecture} Table \ref{table:networkArch} shows the detailed layout of the UBotNet architecture. The three major sections of the architecture are: UNet Encoder, Bottleneck Transformer and UNet Decoder.
\begin{table*}[!ht]
\centering
\caption{\emph{Quantitative results for the comparison between the positional embedding used in the UBotNet architecture for self-attention.} The results for the Relative Positional Embedding are repeated from our main paper for the comparison.} 
\resizebox{\textwidth}{!}{%
\begin{tabular}{@{}llllccclllccc@{}} 
\toprule
\multicolumn{1}{c}{} & \multicolumn{3}{c}{\textbf{Depth Error $\downarrow$}} & \multicolumn{3}{c}{\textbf{Depth Accuracy $\uparrow$}} & \multicolumn{3}{c}{\textbf{Normal Error $\downarrow$}} & \multicolumn{3}{c}{\textbf{Normal Accuracy $\uparrow$}}\\ 

\cmidrule(lr){2-4} \cmidrule(lr){5-7} \cmidrule(lr){8-10} \cmidrule(lr) {11-13}
 
 \multicolumn{1}{c}{\textbf{Method}} & RMSE & MRE & RMSE log & \textbf{$\delta1 < 1.25$} & \textbf{$\delta2  < 1.25^2$} & \textbf{$\delta3  < 1.25^3$}  & Mean & Median & RMSE & \multicolumn{1}{c}{5.0$^\circ$} & \multicolumn{1}{c}{7.5$^\circ$} & \multicolumn{1}{c}{11.25$^\circ$} \\
%  \multicolumn{1}{c}{\textbf{Method}} & \multicolumn{1}{c}{\textbf{No. of parameters}} & Abs Rel & log10 & RMSE log & \textbf{$\delta1 < 1.25$} & \textbf{$\delta2  < 1.25^2$} & \textbf{$\delta3  < 1.25^3$} \\
 \midrule

\multicolumn{1}{c}{Absolute Pos. Emb.}  & \multicolumn{1}{c}{\textbf{0.053}} & \multicolumn{1}{c}{0.290} & \multicolumn{1}{c} {{0.152}} & \multicolumn{1}{c}{{0.691}} & \multicolumn{1}{c}{0.871} & \multicolumn{1}{c}{0.925} & \multicolumn{1}{c}{8.65} & \multicolumn{1}{c}{3.98} & \multicolumn{1}{c} {{13.99}} & \multicolumn{1}{c}{54.26} & \multicolumn{1}{c}{63.00} & \multicolumn{1}{c}{73.23} \\ 

\multicolumn{1}{c}{Relative Pos. Emb.} & \multicolumn{1}{c}{{0.054}} & \multicolumn{1}{c}{\textbf{0.271}} & \multicolumn{1}{c} {\textbf{0.151}} & \multicolumn{1}{c}{\textbf{0.712}} & \multicolumn{1}{c}{\textbf{0.875}} & \multicolumn{1}{c}{\textbf{0.926}} & \multicolumn{1}{c}{\textbf{7.44}} & \multicolumn{1}{c}{\textbf{3.61}} & \multicolumn{1}{c} {\textbf{12.12}} & \multicolumn{1}{c}{\textbf{56.80}} & \multicolumn{1}{c}{\textbf{67.28}} & \multicolumn{1}{c}{\textbf{78.52}}\\ 
\bottomrule
% \multicolumn{1}{c}{} &  &  &  &  &  &  &  \\
\end{tabular}%
}
\label{table:positionalEmbedding}
\end{table*}

\begin{table*}[!ht]
\centering
\caption{\emph{UBotNet Network Architecture.} The network consists of three major segments: UNet Encoder, Bottleneck Transformer and UNet Decoder. Note: To simplify the table we have included the final dense and sigmoid layers in the Decoder towards the end.}
\resizebox{0.8\textwidth}{!}{%
\begin{tabular}{@{}lllllllll@{}}
\toprule
\multicolumn{3}{c}{\textbf{UNet Encoder}} & \multicolumn{3}{c}{\textbf{Bottleneck Transformer}} & \multicolumn{3}{c}{\textbf{UNet Decoder}}\\

\cmidrule(lr){1-3} \cmidrule(lr){4-6} \cmidrule(lr){7-9}
 
 \multicolumn{1}{c}{\textbf{Layer}} & \multicolumn{1}{c}{\textbf{Output Shape}} & \multicolumn{1}{c}{\textbf{Params}} & \multicolumn{1}{c}{\textbf{Layer}} & \multicolumn{1}{c}{\textbf{Output Shape}} & \multicolumn{1}{c}{\textbf{Params}} & \multicolumn{1}{c}{\textbf{Layer}} & \multicolumn{1}{c}{\textbf{Output Shape}} & \multicolumn{1}{c}{\textbf{Params}} \\
%  \multicolumn{1}{c}{\textbf{Method}} & \multicolumn{1}{c}{\textbf{No. of parameters}} & Abs Rel & log10 & RMSE log & \textbf{$\delta1 < 1.25$} & \textbf{$\delta2  < 1.25^2$} & \textbf{$\delta3  < 1.25^3$} \\
 \midrule
\multicolumn{1}{c}{Conv2d-1} & \multicolumn{1}{c}{$128, 512, 256$} & \multicolumn{1}{c}{$3,456$} & \multicolumn{1}{c}{Identity-64} & \multicolumn{1}{c}{$2048, 32, 16$} & \multicolumn{1}{c}{$0$} & \multicolumn{1}{c}{ConvTranspose2d-158} & \multicolumn{1}{c}{$1024, 64, 32$} & \multicolumn{1}{c}{$8,389,632$} \\\multicolumn{1}{c}{BatchNorm2d-2} & \multicolumn{1}{c}{$128, 512, 256$} & \multicolumn{1}{c}{$256$} & \multicolumn{1}{c}{Conv2d-65} & \multicolumn{1}{c}{$512, 32, 16$} & \multicolumn{1}{c}{$1,048,576$} & \multicolumn{1}{c}{Conv2d-159} & \multicolumn{1}{c}{$1024, 64, 32$} & \multicolumn{1}{c}{$18,874,368$} \\\multicolumn{1}{c}{ReLU-3} & \multicolumn{1}{c}{$128, 512, 256$} & \multicolumn{1}{c}{$0$} & \multicolumn{1}{c}{BatchNorm2d-66} & \multicolumn{1}{c}{$512, 32, 16$} & \multicolumn{1}{c}{$1,024$} & \multicolumn{1}{c}{BatchNorm2d-160} & \multicolumn{1}{c}{$1024, 64, 32$} & \multicolumn{1}{c}{$2,048$} \\\multicolumn{1}{c}{Conv2d-4} & \multicolumn{1}{c}{$128, 512, 256$} & \multicolumn{1}{c}{$147,456$} & \multicolumn{1}{c}{ReLU-67} & \multicolumn{1}{c}{$512, 32, 16$} & \multicolumn{1}{c}{$0$} & \multicolumn{1}{c}{ReLU-161} & \multicolumn{1}{c}{$1024, 64, 32$} & \multicolumn{1}{c}{$0$} \\\multicolumn{1}{c}{BatchNorm2d-5} & \multicolumn{1}{c}{$128, 512, 256$} & \multicolumn{1}{c}{$256$} & \multicolumn{1}{c}{ReLU-68} & \multicolumn{1}{c}{$512, 32, 16$} & \multicolumn{1}{c}{$0$} & \multicolumn{1}{c}{Conv2d-162} & \multicolumn{1}{c}{$1024, 64, 32$} & \multicolumn{1}{c}{$9,437,184$} \\\multicolumn{1}{c}{ReLU-6} & \multicolumn{1}{c}{$128, 512, 256$} & \multicolumn{1}{c}{$0$} & \multicolumn{1}{c}{ReLU-69} & \multicolumn{1}{c}{$512, 32, 16$} & \multicolumn{1}{c}{$0$} & \multicolumn{1}{c}{BatchNorm2d-163} & \multicolumn{1}{c}{$1024, 64, 32$} & \multicolumn{1}{c}{$2,048$} \\\multicolumn{1}{c}{DoubleConv-7} & \multicolumn{1}{c}{$128, 512, 256$} & \multicolumn{1}{c}{$0$} & \multicolumn{1}{c}{ReLU-70} & \multicolumn{1}{c}{$512, 32, 16$} & \multicolumn{1}{c}{$0$} & \multicolumn{1}{c}{ReLU-164} & \multicolumn{1}{c}{$1024, 64, 32$} & \multicolumn{1}{c}{$0$} \\\multicolumn{1}{c}{MaxPool2d-8} & \multicolumn{1}{c}{$128, 511, 255$} & \multicolumn{1}{c}{$0$} & \multicolumn{1}{c}{ReLU-71} & \multicolumn{1}{c}{$512, 32, 16$} & \multicolumn{1}{c}{$0$} & \multicolumn{1}{c}{DoubleConv-165} & \multicolumn{1}{c}{$1024, 64, 32$} & \multicolumn{1}{c}{$0$} \\\multicolumn{1}{c}{MaxPool2d-9} & \multicolumn{1}{c}{$128, 511, 255$} & \multicolumn{1}{c}{$0$} & \multicolumn{1}{c}{ReLU-72} & \multicolumn{1}{c}{$512, 32, 16$} & \multicolumn{1}{c}{$0$} & \multicolumn{1}{c}{UNet-up-block-166} & \multicolumn{1}{c}{$1024, 64, 32$} & \multicolumn{1}{c}{$0$} \\\multicolumn{1}{c}{ReflectionPad2d-10} & \multicolumn{1}{c}{$128, 514, 258$} & \multicolumn{1}{c}{$0$} & \multicolumn{1}{c}{Conv2d-73} & \multicolumn{1}{c}{$1024, 32, 16$} & \multicolumn{1}{c}{$524,288$} & \multicolumn{1}{c}{ConvTranspose2d-167} & \multicolumn{1}{c}{$512, 128, 64$} & \multicolumn{1}{c}{$2,097,664$} \\\multicolumn{1}{c}{ReflectionPad2d-11} & \multicolumn{1}{c}{$128, 514, 258$} & \multicolumn{1}{c}{$0$} & \multicolumn{1}{c}{Conv2d-74} & \multicolumn{1}{c}{$512, 32, 16$} & \multicolumn{1}{c}{$262,144$} & \multicolumn{1}{c}{Conv2d-168} & \multicolumn{1}{c}{$512, 128, 64$} & \multicolumn{1}{c}{$4,718,592$} \\\multicolumn{1}{c}{BlurPool-12} & \multicolumn{1}{c}{$128, 256, 128$} & \multicolumn{1}{c}{$0$} & \multicolumn{1}{c}{RelPosEmb-75} & \multicolumn{1}{c}{$4, 512, 512$} & \multicolumn{1}{c}{$0$} & \multicolumn{1}{c}{BatchNorm2d-169} & \multicolumn{1}{c}{$512, 128, 64$} & \multicolumn{1}{c}{$1,024$} \\\multicolumn{1}{c}{BlurPool-13} & \multicolumn{1}{c}{$128, 256, 128$} & \multicolumn{1}{c}{$0$} & \multicolumn{1}{c}{Softmax-76} & \multicolumn{1}{c}{$4, 512, 512$} & \multicolumn{1}{c}{$0$} & \multicolumn{1}{c}{ReLU-170} & \multicolumn{1}{c}{$512, 128, 64$} & \multicolumn{1}{c}{$0$} \\\multicolumn{1}{c}{Conv2d-14} & \multicolumn{1}{c}{$256, 256, 128$} & \multicolumn{1}{c}{$294,912$} & \multicolumn{1}{c}{MHSA-77} & \multicolumn{1}{c}{$512, 32, 16$} & \multicolumn{1}{c}{$0$} & \multicolumn{1}{c}{Conv2d-171} & \multicolumn{1}{c}{$512, 128, 64$} & \multicolumn{1}{c}{$2,359,296$} \\\multicolumn{1}{c}{BatchNorm2d-15} & \multicolumn{1}{c}{$256, 256, 128$} & \multicolumn{1}{c}{$512$} & \multicolumn{1}{c}{Identity-78} & \multicolumn{1}{c}{$512, 32, 16$} & \multicolumn{1}{c}{$0$} & \multicolumn{1}{c}{BatchNorm2d-172} & \multicolumn{1}{c}{$512, 128, 64$} & \multicolumn{1}{c}{$1,024$} \\\multicolumn{1}{c}{ReLU-16} & \multicolumn{1}{c}{$256, 256, 128$} & \multicolumn{1}{c}{$0$} & \multicolumn{1}{c}{BatchNorm2d-79} & \multicolumn{1}{c}{$512, 32, 16$} & \multicolumn{1}{c}{$1,024$} & \multicolumn{1}{c}{ReLU-173} & \multicolumn{1}{c}{$512, 128, 64$} & \multicolumn{1}{c}{$0$} \\\multicolumn{1}{c}{Conv2d-17} & \multicolumn{1}{c}{$256, 256, 128$} & \multicolumn{1}{c}{$589,824$} & \multicolumn{1}{c}{ReLU-80} & \multicolumn{1}{c}{$512, 32, 16$} & \multicolumn{1}{c}{$0$} & \multicolumn{1}{c}{DoubleConv-174} & \multicolumn{1}{c}{$512, 128, 64$} & \multicolumn{1}{c}{$0$} \\\multicolumn{1}{c}{BatchNorm2d-18} & \multicolumn{1}{c}{$256, 256, 128$} & \multicolumn{1}{c}{$512$} & \multicolumn{1}{c}{ReLU-81} & \multicolumn{1}{c}{$512, 32, 16$} & \multicolumn{1}{c}{$0$} & \multicolumn{1}{c}{UNet-up-block-175} & \multicolumn{1}{c}{$512, 128, 64$} & \multicolumn{1}{c}{$0$} \\\multicolumn{1}{c}{ReLU-19} & \multicolumn{1}{c}{$256, 256, 128$} & \multicolumn{1}{c}{$0$} & \multicolumn{1}{c}{ReLU-82} & \multicolumn{1}{c}{$512, 32, 16$} & \multicolumn{1}{c}{$0$} & \multicolumn{1}{c}{ConvTranspose2d-176} & \multicolumn{1}{c}{$256, 256, 128$} & \multicolumn{1}{c}{$524,544$} \\\multicolumn{1}{c}{DoubleConv-20} & \multicolumn{1}{c}{$256, 256, 128$} & \multicolumn{1}{c}{$0$} & \multicolumn{1}{c}{ReLU-83} & \multicolumn{1}{c}{$512, 32, 16$} & \multicolumn{1}{c}{$0$} & \multicolumn{1}{c}{Conv2d-177} & \multicolumn{1}{c}{$256, 256, 128$} & \multicolumn{1}{c}{$1,179,648$} \\\multicolumn{1}{c}{UNet-down-block-21} & \multicolumn{1}{c}{$256, 256, 128$} & \multicolumn{1}{c}{$0$} & \multicolumn{1}{c}{ReLU-84} & \multicolumn{1}{c}{$512, 32, 16$} & \multicolumn{1}{c}{$0$} & \multicolumn{1}{c}{BatchNorm2d-178} & \multicolumn{1}{c}{$256, 256, 128$} & \multicolumn{1}{c}{$512$} \\\multicolumn{1}{c}{MaxPool2d-22} & \multicolumn{1}{c}{$256, 255, 127$} & \multicolumn{1}{c}{$0$} & \multicolumn{1}{c}{ReLU-85} & \multicolumn{1}{c}{$512, 32, 16$} & \multicolumn{1}{c}{$0$} & \multicolumn{1}{c}{ReLU-179} & \multicolumn{1}{c}{$256, 256, 128$} & \multicolumn{1}{c}{$0$} \\\multicolumn{1}{c}{MaxPool2d-23} & \multicolumn{1}{c}{$256, 255, 127$} & \multicolumn{1}{c}{$0$} & \multicolumn{1}{c}{Conv2d-86} & \multicolumn{1}{c}{$2048, 32, 16$} & \multicolumn{1}{c}{$1,048,576$} & \multicolumn{1}{c}{Conv2d-180} & \multicolumn{1}{c}{$256, 256, 128$} & \multicolumn{1}{c}{$589,824$} \\\multicolumn{1}{c}{ReflectionPad2d-24} & \multicolumn{1}{c}{$256, 258, 130$} & \multicolumn{1}{c}{$0$} & \multicolumn{1}{c}{BatchNorm2d-87} & \multicolumn{1}{c}{$2048, 32, 16$} & \multicolumn{1}{c}{$4,096$} & \multicolumn{1}{c}{BatchNorm2d-181} & \multicolumn{1}{c}{$256, 256, 128$} & \multicolumn{1}{c}{$512$} \\\multicolumn{1}{c}{ReflectionPad2d-25} & \multicolumn{1}{c}{$256, 258, 130$} & \multicolumn{1}{c}{$0$} & \multicolumn{1}{c}{ReLU-88} & \multicolumn{1}{c}{$2048, 32, 16$} & \multicolumn{1}{c}{$0$} & \multicolumn{1}{c}{ReLU-182} & \multicolumn{1}{c}{$256, 256, 128$} & \multicolumn{1}{c}{$0$} \\\multicolumn{1}{c}{BlurPool-26} & \multicolumn{1}{c}{$256, 128, 64$} & \multicolumn{1}{c}{$0$} & \multicolumn{1}{c}{ReLU-89} & \multicolumn{1}{c}{$2048, 32, 16$} & \multicolumn{1}{c}{$0$} & \multicolumn{1}{c}{DoubleConv-183} & \multicolumn{1}{c}{$256, 256, 128$} & \multicolumn{1}{c}{$0$} \\\multicolumn{1}{c}{BlurPool-27} & \multicolumn{1}{c}{$256, 128, 64$} & \multicolumn{1}{c}{$0$} & \multicolumn{1}{c}{ReLU-90} & \multicolumn{1}{c}{$2048, 32, 16$} & \multicolumn{1}{c}{$0$} & \multicolumn{1}{c}{UNet-up-block-184} & \multicolumn{1}{c}{$256, 256, 128$} & \multicolumn{1}{c}{$0$} \\\multicolumn{1}{c}{Conv2d-28} & \multicolumn{1}{c}{$512, 128, 64$} & \multicolumn{1}{c}{$1,179,648$} & \multicolumn{1}{c}{ReLU-91} & \multicolumn{1}{c}{$2048, 32, 16$} & \multicolumn{1}{c}{$0$} & \multicolumn{1}{c}{ConvTranspose2d-185} & \multicolumn{1}{c}{$128, 512, 256$} & \multicolumn{1}{c}{$131,200$} \\\multicolumn{1}{c}{BatchNorm2d-29} & \multicolumn{1}{c}{$512, 128, 64$} & \multicolumn{1}{c}{$1,024$} & \multicolumn{1}{c}{ReLU-92} & \multicolumn{1}{c}{$2048, 32, 16$} & \multicolumn{1}{c}{$0$} & \multicolumn{1}{c}{Conv2d-186} & \multicolumn{1}{c}{$128, 512, 256$} & \multicolumn{1}{c}{$294,912$} \\\multicolumn{1}{c}{ReLU-30} & \multicolumn{1}{c}{$512, 128, 64$} & \multicolumn{1}{c}{$0$} & \multicolumn{1}{c}{ReLU-93} & \multicolumn{1}{c}{$2048, 32, 16$} & \multicolumn{1}{c}{$0$} & \multicolumn{1}{c}{BatchNorm2d-187} & \multicolumn{1}{c}{$128, 512, 256$} & \multicolumn{1}{c}{$256$} \\\multicolumn{1}{c}{Conv2d-31} & \multicolumn{1}{c}{$512, 128, 64$} & \multicolumn{1}{c}{$2,359,296$} & \multicolumn{1}{c}{BoTBlock-94} & \multicolumn{1}{c}{$2048, 32, 16$} & \multicolumn{1}{c}{$0$} & \multicolumn{1}{c}{ReLU-188} & \multicolumn{1}{c}{$128, 512, 256$} & \multicolumn{1}{c}{$0$} \\\multicolumn{1}{c}{BatchNorm2d-32} & \multicolumn{1}{c}{$512, 128, 64$} & \multicolumn{1}{c}{$1,024$} & \multicolumn{1}{c}{Identity-95} & \multicolumn{1}{c}{$2048, 32, 16$} & \multicolumn{1}{c}{$0$} & \multicolumn{1}{c}{Conv2d-189} & \multicolumn{1}{c}{$128, 512, 256$} & \multicolumn{1}{c}{$147,456$} \\\multicolumn{1}{c}{ReLU-33} & \multicolumn{1}{c}{$512, 128, 64$} & \multicolumn{1}{c}{$0$} & \multicolumn{1}{c}{Conv2d-96} & \multicolumn{1}{c}{$512, 32, 16$} & \multicolumn{1}{c}{$1,048,576$} & \multicolumn{1}{c}{BatchNorm2d-190} & \multicolumn{1}{c}{$128, 512, 256$} & \multicolumn{1}{c}{$256$} \\\multicolumn{1}{c}{DoubleConv-34} & \multicolumn{1}{c}{$512, 128, 64$} & \multicolumn{1}{c}{$0$} & \multicolumn{1}{c}{BatchNorm2d-97} & \multicolumn{1}{c}{$512, 32, 16$} & \multicolumn{1}{c}{$1,024$} & \multicolumn{1}{c}{ReLU-191} & \multicolumn{1}{c}{$128, 512, 256$} & \multicolumn{1}{c}{$0$} \\\multicolumn{1}{c}{UNet-down-block-35} & \multicolumn{1}{c}{$512, 128, 64$} & \multicolumn{1}{c}{$0$} & \multicolumn{1}{c}{ReLU-98} & \multicolumn{1}{c}{$512, 32, 16$} & \multicolumn{1}{c}{$0$} & \multicolumn{1}{c}{DoubleConv-192} & \multicolumn{1}{c}{$128, 512, 256$} & \multicolumn{1}{c}{$0$} \\\multicolumn{1}{c}{MaxPool2d-36} & \multicolumn{1}{c}{$512, 127, 63$} & \multicolumn{1}{c}{$0$} & \multicolumn{1}{c}{ReLU-99} & \multicolumn{1}{c}{$512, 32, 16$} & \multicolumn{1}{c}{$0$} & \multicolumn{1}{c}{UNet-up-block-193} & \multicolumn{1}{c}{$128, 512, 256$} & \multicolumn{1}{c}{$0$} \\\multicolumn{1}{c}{MaxPool2d-37} & \multicolumn{1}{c}{$512, 127, 63$} & \multicolumn{1}{c}{$0$} & \multicolumn{1}{c}{ReLU-100} & \multicolumn{1}{c}{$512, 32, 16$} & \multicolumn{1}{c}{$0$} & \multicolumn{1}{c}{Linear-194} & \multicolumn{1}{c}{$512, 256, 512$} & \multicolumn{1}{c}{$66,048$} \\\multicolumn{1}{c}{ReflectionPad2d-38} & \multicolumn{1}{c}{$512, 130, 66$} & \multicolumn{1}{c}{$0$} & \multicolumn{1}{c}{ReLU-101} & \multicolumn{1}{c}{$512, 32, 16$} & \multicolumn{1}{c}{$0$} & \multicolumn{1}{c}{ReLU-195} & \multicolumn{1}{c}{$512, 256, 512$} & \multicolumn{1}{c}{$0$} \\\multicolumn{1}{c}{ReflectionPad2d-39} & \multicolumn{1}{c}{$512, 130, 66$} & \multicolumn{1}{c}{$0$} & \multicolumn{1}{c}{ReLU-102} & \multicolumn{1}{c}{$512, 32, 16$} & \multicolumn{1}{c}{$0$} & \multicolumn{1}{c}{Dropout-196} & \multicolumn{1}{c}{$512, 256, 512$} & \multicolumn{1}{c}{$0$} \\\multicolumn{1}{c}{BlurPool-40} & \multicolumn{1}{c}{$512, 64, 32$} & \multicolumn{1}{c}{$0$} & \multicolumn{1}{c}{ReLU-103} & \multicolumn{1}{c}{$512, 32, 16$} & \multicolumn{1}{c}{$0$} & \multicolumn{1}{c}{Linear-197} & \multicolumn{1}{c}{$512, 256, 128$} & \multicolumn{1}{c}{$65,664$} \\\multicolumn{1}{c}{BlurPool-41} & \multicolumn{1}{c}{$512, 64, 32$} & \multicolumn{1}{c}{$0$} & \multicolumn{1}{c}{Conv2d-104} & \multicolumn{1}{c}{$1024, 32, 16$} & \multicolumn{1}{c}{$524,288$} & \multicolumn{1}{c}{ReLU-198} & \multicolumn{1}{c}{$512, 256, 128$} & \multicolumn{1}{c}{$0$} \\\multicolumn{1}{c}{Conv2d-42} & \multicolumn{1}{c}{$1024, 64, 32$} & \multicolumn{1}{c}{$4,718,592$} & \multicolumn{1}{c}{Conv2d-105} & \multicolumn{1}{c}{$512, 32, 16$} & \multicolumn{1}{c}{$262,144$} & \multicolumn{1}{c}{Dropout-199} & \multicolumn{1}{c}{$512, 256, 128$} & \multicolumn{1}{c}{$0$} \\\multicolumn{1}{c}{BatchNorm2d-43} & \multicolumn{1}{c}{$1024, 64, 32$} & \multicolumn{1}{c}{$2,048$} & \multicolumn{1}{c}{RelPosEmb-106} & \multicolumn{1}{c}{$4, 512, 512$} & \multicolumn{1}{c}{$0$} & \multicolumn{1}{c}{Linear-200} & \multicolumn{1}{c}{$512, 256, 1$} & \multicolumn{1}{c}{$129$} \\\multicolumn{1}{c}{ReLU-44} & \multicolumn{1}{c}{$1024, 64, 32$} & \multicolumn{1}{c}{$0$} & \multicolumn{1}{c}{Softmax-107} & \multicolumn{1}{c}{$4, 512, 512$} & \multicolumn{1}{c}{$0$} & \multicolumn{1}{c}{Sigmoid-201} & \multicolumn{1}{c}{$512, 256, 1$} & \multicolumn{1}{c}{$0$} \\\multicolumn{1}{c}{Conv2d-45} & \multicolumn{1}{c}{$1024, 64, 32$} & \multicolumn{1}{c}{$9,437,184$} & \multicolumn{1}{c}{MHSA-108} & \multicolumn{1}{c}{$512, 32, 16$} & \multicolumn{1}{c}{$0$} & \multicolumn{1}{c}{Linear-202} & \multicolumn{1}{c}{$512, 256, 3$} & \multicolumn{1}{c}{$387$} \\\multicolumn{1}{c}{BatchNorm2d-46} & \multicolumn{1}{c}{$1024, 64, 32$} & \multicolumn{1}{c}{$2,048$} & \multicolumn{1}{c}{Identity-109} & \multicolumn{1}{c}{$512, 32, 16$} & \multicolumn{1}{c}{$0$} & \multicolumn{1}{c}{Sigmoid-203} & \multicolumn{1}{c}{$512, 256, 3$} & \multicolumn{1}{c}{$0$} \\\multicolumn{1}{c}{ReLU-47} & \multicolumn{1}{c}{$1024, 64, 32$} & \multicolumn{1}{c}{$0$} & \multicolumn{1}{c}{BatchNorm2d-110} & \multicolumn{1}{c}{$512, 32, 16$} & \multicolumn{1}{c}{$1,024$} & \multicolumn{1}{c}{ } & \multicolumn{1}{c}{ } & \multicolumn{1}{c}{ } \\\multicolumn{1}{c}{DoubleConv-48} & \multicolumn{1}{c}{$1024, 64, 32$} & \multicolumn{1}{c}{$0$} & \multicolumn{1}{c}{ReLU-111} & \multicolumn{1}{c}{$512, 32, 16$} & \multicolumn{1}{c}{$0$} & \multicolumn{1}{c}{ } & \multicolumn{1}{c}{ } & \multicolumn{1}{c}{ } \\\multicolumn{1}{c}{UNet-down-block-49} & \multicolumn{1}{c}{$1024, 64, 32$} & \multicolumn{1}{c}{$0$} & \multicolumn{1}{c}{ReLU-112} & \multicolumn{1}{c}{$512, 32, 16$} & \multicolumn{1}{c}{$0$} & \multicolumn{1}{c}{ } & \multicolumn{1}{c}{ } & \multicolumn{1}{c}{ } \\\multicolumn{1}{c}{MaxPool2d-50} & \multicolumn{1}{c}{$1024, 63, 31$} & \multicolumn{1}{c}{$0$} & \multicolumn{1}{c}{ReLU-113} & \multicolumn{1}{c}{$512, 32, 16$} & \multicolumn{1}{c}{$0$} & \multicolumn{1}{c}{ } & \multicolumn{1}{c}{ } & \multicolumn{1}{c}{ } \\\multicolumn{1}{c}{MaxPool2d-51} & \multicolumn{1}{c}{$1024, 63, 31$} & \multicolumn{1}{c}{$0$} & \multicolumn{1}{c}{ReLU-114} & \multicolumn{1}{c}{$512, 32, 16$} & \multicolumn{1}{c}{$0$} & \multicolumn{1}{c}{ } & \multicolumn{1}{c}{ } & \multicolumn{1}{c}{ } \\\multicolumn{1}{c}{ReflectionPad2d-52} & \multicolumn{1}{c}{$1024, 66, 34$} & \multicolumn{1}{c}{$0$} & \multicolumn{1}{c}{ReLU-115} & \multicolumn{1}{c}{$512, 32, 16$} & \multicolumn{1}{c}{$0$} & \multicolumn{1}{c}{ } & \multicolumn{1}{c}{ } & \multicolumn{1}{c}{ } \\\multicolumn{1}{c}{ReflectionPad2d-53} & \multicolumn{1}{c}{$1024, 66, 34$} & \multicolumn{1}{c}{$0$} & \multicolumn{1}{c}{ReLU-116} & \multicolumn{1}{c}{$512, 32, 16$} & \multicolumn{1}{c}{$0$} & \multicolumn{1}{c}{ } & \multicolumn{1}{c}{ } & \multicolumn{1}{c}{ } \\\multicolumn{1}{c}{BlurPool-54} & \multicolumn{1}{c}{$1024, 32, 16$} & \multicolumn{1}{c}{$0$} & \multicolumn{1}{c}{Conv2d-117} & \multicolumn{1}{c}{$2048, 32, 16$} & \multicolumn{1}{c}{$1,048,576$} & \multicolumn{1}{c}{ } & \multicolumn{1}{c}{ } & \multicolumn{1}{c}{ } \\\multicolumn{1}{c}{BlurPool-55} & \multicolumn{1}{c}{$1024, 32, 16$} & \multicolumn{1}{c}{$0$} & \multicolumn{1}{c}{BatchNorm2d-118} & \multicolumn{1}{c}{$2048, 32, 16$} & \multicolumn{1}{c}{$4,096$} & \multicolumn{1}{c}{ } & \multicolumn{1}{c}{ } & \multicolumn{1}{c}{ } \\\multicolumn{1}{c}{Conv2d-56} & \multicolumn{1}{c}{$2048, 32, 16$} & \multicolumn{1}{c}{$18,874,368$} & \multicolumn{1}{c}{ReLU-119} & \multicolumn{1}{c}{$2048, 32, 16$} & \multicolumn{1}{c}{$0$} & \multicolumn{1}{c}{ } & \multicolumn{1}{c}{ } & \multicolumn{1}{c}{ } \\\multicolumn{1}{c}{BatchNorm2d-57} & \multicolumn{1}{c}{$2048, 32, 16$} & \multicolumn{1}{c}{$4,096$} & \multicolumn{1}{c}{ReLU-120} & \multicolumn{1}{c}{$2048, 32, 16$} & \multicolumn{1}{c}{$0$} & \multicolumn{1}{c}{ } & \multicolumn{1}{c}{ } & \multicolumn{1}{c}{ } \\\multicolumn{1}{c}{ReLU-58} & \multicolumn{1}{c}{$2048, 32, 16$} & \multicolumn{1}{c}{$0$} & \multicolumn{1}{c}{ReLU-121} & \multicolumn{1}{c}{$2048, 32, 16$} & \multicolumn{1}{c}{$0$} & \multicolumn{1}{c}{ } & \multicolumn{1}{c}{ } & \multicolumn{1}{c}{ } \\\multicolumn{1}{c}{Conv2d-59} & \multicolumn{1}{c}{$2048, 32, 16$} & \multicolumn{1}{c}{$37,748,736$} & \multicolumn{1}{c}{ReLU-122} & \multicolumn{1}{c}{$2048, 32, 16$} & \multicolumn{1}{c}{$0$} & \multicolumn{1}{c}{ } & \multicolumn{1}{c}{ } & \multicolumn{1}{c}{ } \\\multicolumn{1}{c}{BatchNorm2d-60} & \multicolumn{1}{c}{$2048, 32, 16$} & \multicolumn{1}{c}{$4,096$} & \multicolumn{1}{c}{ReLU-123} & \multicolumn{1}{c}{$2048, 32, 16$} & \multicolumn{1}{c}{$0$} & \multicolumn{1}{c}{ } & \multicolumn{1}{c}{ } & \multicolumn{1}{c}{ } \\\multicolumn{1}{c}{ReLU-61} & \multicolumn{1}{c}{$2048, 32, 16$} & \multicolumn{1}{c}{$0$} & \multicolumn{1}{c}{ReLU-124} & \multicolumn{1}{c}{$2048, 32, 16$} & \multicolumn{1}{c}{$0$} & \multicolumn{1}{c}{ } & \multicolumn{1}{c}{ } & \multicolumn{1}{c}{ } \\\multicolumn{1}{c}{DoubleConv-62} & \multicolumn{1}{c}{$2048, 32, 16$} & \multicolumn{1}{c}{$0$} & \multicolumn{1}{c}{BoTBlock-125} & \multicolumn{1}{c}{$2048, 32, 16$} & \multicolumn{1}{c}{$0$} & \multicolumn{1}{c}{ } & \multicolumn{1}{c}{ } & \multicolumn{1}{c}{ } \\\multicolumn{1}{c}{UNet-down-block-63} & \multicolumn{1}{c}{$2048, 32, 16$} & \multicolumn{1}{c}{$0$} & \multicolumn{1}{c}{Identity-126} & \multicolumn{1}{c}{$2048, 32, 16$} & \multicolumn{1}{c}{$0$} & \multicolumn{1}{c}{ } & \multicolumn{1}{c}{ } & \multicolumn{1}{c}{ } \\\multicolumn{1}{c}{ } & \multicolumn{1}{c}{ } & \multicolumn{1}{c}{ } & \multicolumn{1}{c}{Conv2d-127} & \multicolumn{1}{c}{$512, 32, 16$} & \multicolumn{1}{c}{$1,048,576$} & \multicolumn{1}{c}{ } & \multicolumn{1}{c}{ } & \multicolumn{1}{c}{ } \\\multicolumn{1}{c}{ } & \multicolumn{1}{c}{ } & \multicolumn{1}{c}{ } & \multicolumn{1}{c}{BatchNorm2d-128} & \multicolumn{1}{c}{$512, 32, 16$} & \multicolumn{1}{c}{$1,024$} & \multicolumn{1}{c}{ } & \multicolumn{1}{c}{ } & \multicolumn{1}{c}{ } \\\multicolumn{1}{c}{ } & \multicolumn{1}{c}{ } & \multicolumn{1}{c}{ } & \multicolumn{1}{c}{ReLU-129} & \multicolumn{1}{c}{$512, 32, 16$} & \multicolumn{1}{c}{$0$} & \multicolumn{1}{c}{ } & \multicolumn{1}{c}{ } & \multicolumn{1}{c}{ } \\\multicolumn{1}{c}{ } & \multicolumn{1}{c}{ } & \multicolumn{1}{c}{ } & \multicolumn{1}{c}{ReLU-130} & \multicolumn{1}{c}{$512, 32, 16$} & \multicolumn{1}{c}{$0$} & \multicolumn{1}{c}{ } & \multicolumn{1}{c}{ } & \multicolumn{1}{c}{ } \\\multicolumn{1}{c}{ } & \multicolumn{1}{c}{ } & \multicolumn{1}{c}{ } & \multicolumn{1}{c}{ReLU-131} & \multicolumn{1}{c}{$512, 32, 16$} & \multicolumn{1}{c}{$0$} & \multicolumn{1}{c}{ } & \multicolumn{1}{c}{ } & \multicolumn{1}{c}{ } \\\multicolumn{1}{c}{ } & \multicolumn{1}{c}{ } & \multicolumn{1}{c}{ } & \multicolumn{1}{c}{ReLU-132} & \multicolumn{1}{c}{$512, 32, 16$} & \multicolumn{1}{c}{$0$} & \multicolumn{1}{c}{ } & \multicolumn{1}{c}{ } & \multicolumn{1}{c}{ } \\\multicolumn{1}{c}{ } & \multicolumn{1}{c}{ } & \multicolumn{1}{c}{ } & \multicolumn{1}{c}{ReLU-133} & \multicolumn{1}{c}{$512, 32, 16$} & \multicolumn{1}{c}{$0$} & \multicolumn{1}{c}{ } & \multicolumn{1}{c}{ } & \multicolumn{1}{c}{ } \\\multicolumn{1}{c}{ } & \multicolumn{1}{c}{ } & \multicolumn{1}{c}{ } & \multicolumn{1}{c}{ReLU-134} & \multicolumn{1}{c}{$512, 32, 16$} & \multicolumn{1}{c}{$0$} & \multicolumn{1}{c}{ } & \multicolumn{1}{c}{ } & \multicolumn{1}{c}{ } \\\multicolumn{1}{c}{ } & \multicolumn{1}{c}{ } & \multicolumn{1}{c}{ } & \multicolumn{1}{c}{Conv2d-135} & \multicolumn{1}{c}{$1024, 32, 16$} & \multicolumn{1}{c}{$524,288$} & \multicolumn{1}{c}{ } & \multicolumn{1}{c}{ } & \multicolumn{1}{c}{ } \\\multicolumn{1}{c}{ } & \multicolumn{1}{c}{ } & \multicolumn{1}{c}{ } & \multicolumn{1}{c}{Conv2d-136} & \multicolumn{1}{c}{$512, 32, 16$} & \multicolumn{1}{c}{$262,144$} & \multicolumn{1}{c}{ } & \multicolumn{1}{c}{ } & \multicolumn{1}{c}{ } \\\multicolumn{1}{c}{ } & \multicolumn{1}{c}{ } & \multicolumn{1}{c}{ } & \multicolumn{1}{c}{RelPosEmb-137} & \multicolumn{1}{c}{$4, 512, 512$} & \multicolumn{1}{c}{$0$} & \multicolumn{1}{c}{ } & \multicolumn{1}{c}{ } & \multicolumn{1}{c}{ } \\\multicolumn{1}{c}{ } & \multicolumn{1}{c}{ } & \multicolumn{1}{c}{ } & \multicolumn{1}{c}{Softmax-138} & \multicolumn{1}{c}{$4, 512, 512$} & \multicolumn{1}{c}{$0$} & \multicolumn{1}{c}{ } & \multicolumn{1}{c}{ } & \multicolumn{1}{c}{ } \\\multicolumn{1}{c}{ } & \multicolumn{1}{c}{ } & \multicolumn{1}{c}{ } & \multicolumn{1}{c}{MHSA-139} & \multicolumn{1}{c}{$512, 32, 16$} & \multicolumn{1}{c}{$0$} & \multicolumn{1}{c}{ } & \multicolumn{1}{c}{ } & \multicolumn{1}{c}{ } \\\multicolumn{1}{c}{ } & \multicolumn{1}{c}{ } & \multicolumn{1}{c}{ } & \multicolumn{1}{c}{Identity-140} & \multicolumn{1}{c}{$512, 32, 16$} & \multicolumn{1}{c}{$0$} & \multicolumn{1}{c}{ } & \multicolumn{1}{c}{ } & \multicolumn{1}{c}{ } \\\multicolumn{1}{c}{ } & \multicolumn{1}{c}{ } & \multicolumn{1}{c}{ } & \multicolumn{1}{c}{BatchNorm2d-141} & \multicolumn{1}{c}{$512, 32, 16$} & \multicolumn{1}{c}{$1,024$} & \multicolumn{1}{c}{ } & \multicolumn{1}{c}{ } & \multicolumn{1}{c}{ } \\\multicolumn{1}{c}{ } & \multicolumn{1}{c}{ } & \multicolumn{1}{c}{ } & \multicolumn{1}{c}{ReLU-142} & \multicolumn{1}{c}{$512, 32, 16$} & \multicolumn{1}{c}{$0$} & \multicolumn{1}{c}{ } & \multicolumn{1}{c}{ } & \multicolumn{1}{c}{ } \\\multicolumn{1}{c}{ } & \multicolumn{1}{c}{ } & \multicolumn{1}{c}{ } & \multicolumn{1}{c}{ReLU-143} & \multicolumn{1}{c}{$512, 32, 16$} & \multicolumn{1}{c}{$0$} & \multicolumn{1}{c}{ } & \multicolumn{1}{c}{ } & \multicolumn{1}{c}{ } \\\multicolumn{1}{c}{ } & \multicolumn{1}{c}{ } & \multicolumn{1}{c}{ } & \multicolumn{1}{c}{ReLU-144} & \multicolumn{1}{c}{$512, 32, 16$} & \multicolumn{1}{c}{$0$} & \multicolumn{1}{c}{ } & \multicolumn{1}{c}{ } & \multicolumn{1}{c}{ } \\\multicolumn{1}{c}{ } & \multicolumn{1}{c}{ } & \multicolumn{1}{c}{ } & \multicolumn{1}{c}{ReLU-145} & \multicolumn{1}{c}{$512, 32, 16$} & \multicolumn{1}{c}{$0$} & \multicolumn{1}{c}{ } & \multicolumn{1}{c}{ } & \multicolumn{1}{c}{ } \\\multicolumn{1}{c}{ } & \multicolumn{1}{c}{ } & \multicolumn{1}{c}{ } & \multicolumn{1}{c}{ReLU-146} & \multicolumn{1}{c}{$512, 32, 16$} & \multicolumn{1}{c}{$0$} & \multicolumn{1}{c}{ } & \multicolumn{1}{c}{ } & \multicolumn{1}{c}{ } \\\multicolumn{1}{c}{ } & \multicolumn{1}{c}{ } & \multicolumn{1}{c}{ } & \multicolumn{1}{c}{ReLU-147} & \multicolumn{1}{c}{$512, 32, 16$} & \multicolumn{1}{c}{$0$} & \multicolumn{1}{c}{ } & \multicolumn{1}{c}{ } & \multicolumn{1}{c}{ } \\\multicolumn{1}{c}{ } & \multicolumn{1}{c}{ } & \multicolumn{1}{c}{ } & \multicolumn{1}{c}{Conv2d-148} & \multicolumn{1}{c}{$2048, 32, 16$} & \multicolumn{1}{c}{$1,048,576$} & \multicolumn{1}{c}{ } & \multicolumn{1}{c}{ } & \multicolumn{1}{c}{ } \\\multicolumn{1}{c}{ } & \multicolumn{1}{c}{ } & \multicolumn{1}{c}{ } & \multicolumn{1}{c}{BatchNorm2d-149} & \multicolumn{1}{c}{$2048, 32, 16$} & \multicolumn{1}{c}{$4,096$} & \multicolumn{1}{c}{ } & \multicolumn{1}{c}{ } & \multicolumn{1}{c}{ } \\\multicolumn{1}{c}{ } & \multicolumn{1}{c}{ } & \multicolumn{1}{c}{ } & \multicolumn{1}{c}{ReLU-150} & \multicolumn{1}{c}{$2048, 32, 16$} & \multicolumn{1}{c}{$0$} & \multicolumn{1}{c}{ } & \multicolumn{1}{c}{ } & \multicolumn{1}{c}{ } \\\multicolumn{1}{c}{ } & \multicolumn{1}{c}{ } & \multicolumn{1}{c}{ } & \multicolumn{1}{c}{ReLU-151} & \multicolumn{1}{c}{$2048, 32, 16$} & \multicolumn{1}{c}{$0$} & \multicolumn{1}{c}{ } & \multicolumn{1}{c}{ } & \multicolumn{1}{c}{ } \\\multicolumn{1}{c}{ } & \multicolumn{1}{c}{ } & \multicolumn{1}{c}{ } & \multicolumn{1}{c}{ReLU-152} & \multicolumn{1}{c}{$2048, 32, 16$} & \multicolumn{1}{c}{$0$} & \multicolumn{1}{c}{ } & \multicolumn{1}{c}{ } & \multicolumn{1}{c}{ } \\\multicolumn{1}{c}{ } & \multicolumn{1}{c}{ } & \multicolumn{1}{c}{ } & \multicolumn{1}{c}{ReLU-153} & \multicolumn{1}{c}{$2048, 32, 16$} & \multicolumn{1}{c}{$0$} & \multicolumn{1}{c}{ } & \multicolumn{1}{c}{ } & \multicolumn{1}{c}{ } \\\multicolumn{1}{c}{ } & \multicolumn{1}{c}{ } & \multicolumn{1}{c}{ } & \multicolumn{1}{c}{ReLU-154} & \multicolumn{1}{c}{$2048, 32, 16$} & \multicolumn{1}{c}{$0$} & \multicolumn{1}{c}{ } & \multicolumn{1}{c}{ } & \multicolumn{1}{c}{ } \\\multicolumn{1}{c}{ } & \multicolumn{1}{c}{ } & \multicolumn{1}{c}{ } & \multicolumn{1}{c}{ReLU-155} & \multicolumn{1}{c}{$2048, 32, 16$} & \multicolumn{1}{c}{$0$} & \multicolumn{1}{c}{ } & \multicolumn{1}{c}{ } & \multicolumn{1}{c}{ } \\\multicolumn{1}{c}{ } & \multicolumn{1}{c}{ } & \multicolumn{1}{c}{ } & \multicolumn{1}{c}{BoTBlock-156} & \multicolumn{1}{c}{$2048, 32, 16$} & \multicolumn{1}{c}{$0$} & \multicolumn{1}{c}{ } & \multicolumn{1}{c}{ } & \multicolumn{1}{c}{ } \\\multicolumn{1}{c}{ } & \multicolumn{1}{c}{ } & \multicolumn{1}{c}{ } & \multicolumn{1}{c}{BoTStack-157} & \multicolumn{1}{c}{$2048, 32, 16$} & \multicolumn{1}{c}{$0$} & \multicolumn{1}{c}{ } & \multicolumn{1}{c}{ } & \multicolumn{1}{c}{ } \\

\bottomrule
% \multicolumn{1}{c}{} &  &  &  &  &  &  &  \\
\end{tabular}%
}
\label{table:networkArch}
\end{table*}

\begin{figure*} [!h]
    \centering
    \includegraphics[width=0.85\textwidth]{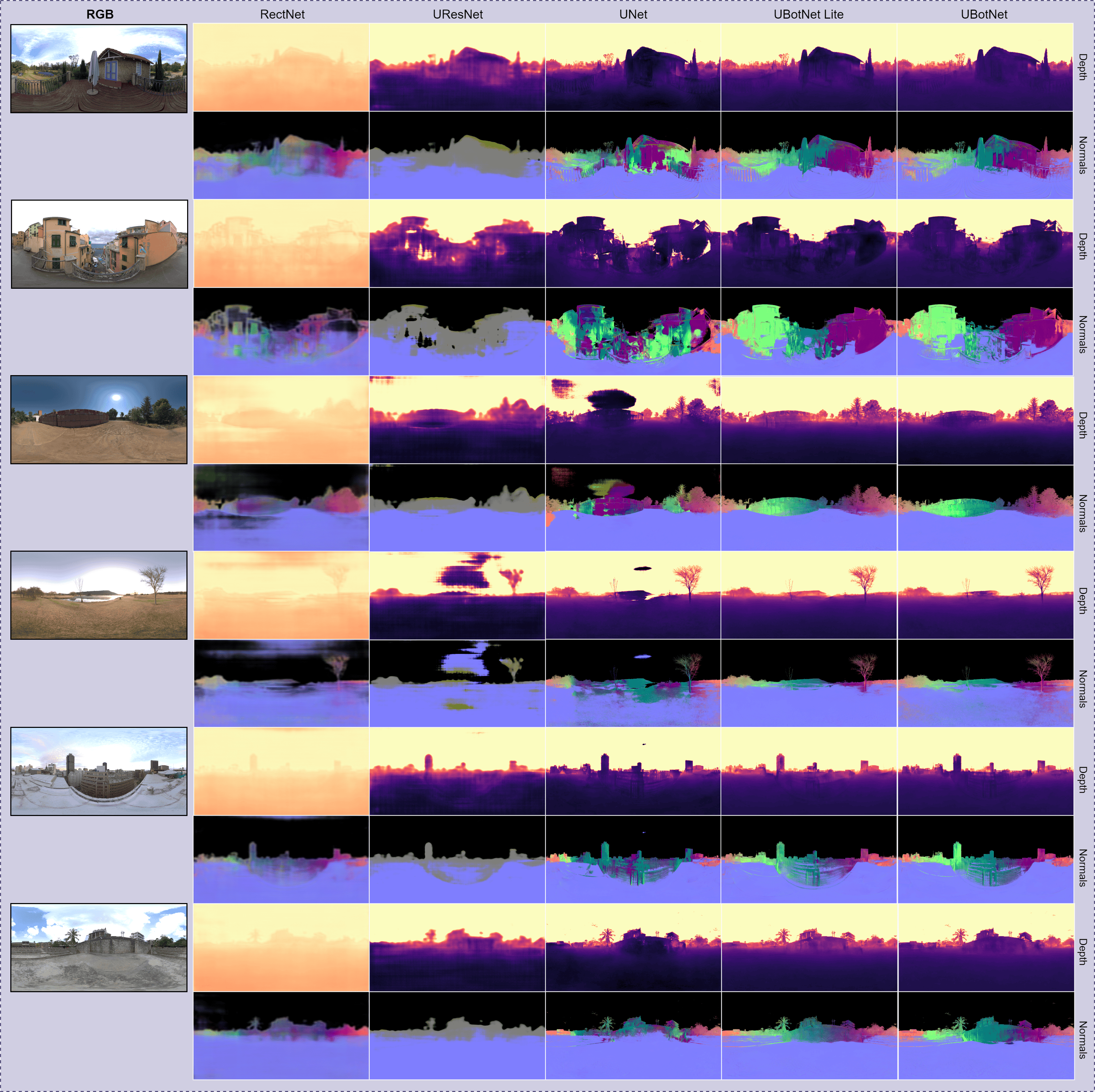}
    \caption{\emph{Depth and Normal estimation on real-world images in the wild.} Comparison between all the networks discussed in main paper for depth and normal estimation on real world images.}
    \label{fig:realWorldAdditionalResults}
\end{figure*}

\begin{figure*} [!h]
    \centering
    \includegraphics[width=\textwidth]{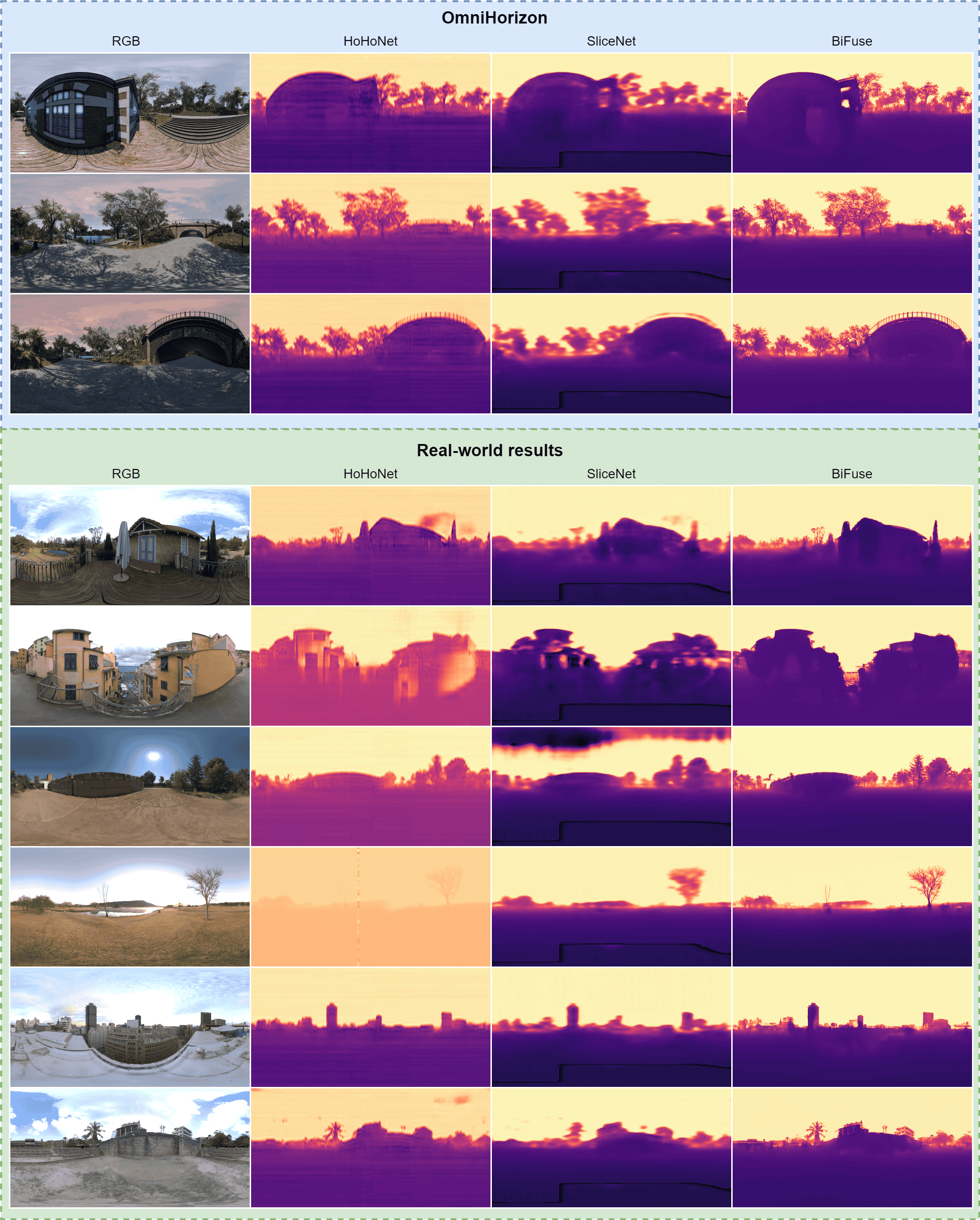}
    \caption{\emph{Qualitative results for monocular depth estimation by \cite{sun2021hohonet}, \cite{Pintore2021SliceNet} and \cite{wang2020bifuse} on OmniHorizon and real-world images.}}
    \label{fig:realWorldAdditionalResults2}
\end{figure*}

\begin{figure*} [!h]
    \centering
    \includegraphics[width=\textwidth]{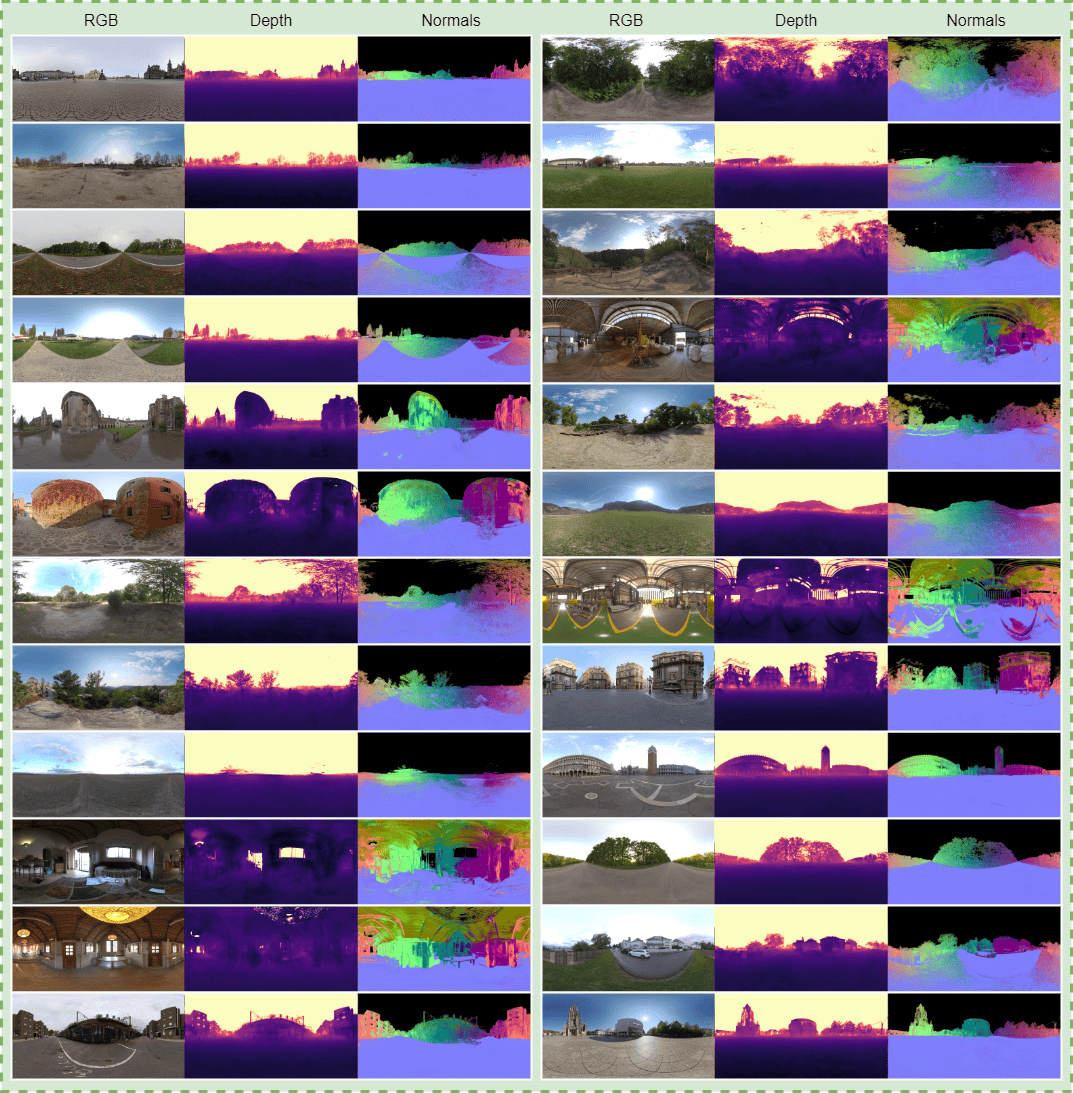}
    \caption{\emph{Examples of depth and normal estimation using UBotNet on real-world images in the wild.}}
    \label{fig:realWorldAdditionalResults3}
\end{figure*}

\section{Addition Results}

Figure~\ref{fig:realWorldAdditionalResults} shows the results for the networks discussed in main paper for depth and normal estimation on 360 images captured from real-world locations. As evident from the results, other methods struggle when estimating the normal information. On the other hand, UBotNet leverages this information and estimates both depth and normal information with increased accuracy. Note that some of networks also struggle with changing sky conditions and hence produce artifacts in those regions. UBotnet is more robust to diverse outdoor lighting and sky conditions. Interestingly, other networks also fail to identify vertical structures (as shown in last image of Figure~\ref{fig:realWorldAdditionalResults}) whereas both UBotNet and UbotNet Lite are able to segment ground from the walls and boundaries.

Figure~\ref{fig:realWorldAdditionalResults2} demonstrates the qualitative results for the architectures that perform only depth estimation. The visual output of depth estimation from the Bifuse \cite{wang2020bifuse} support the quantitative results in the main paper where Bifuse performs really well in OmniHorizon benchmark.
Figure~\ref{fig:realWorldAdditionalResults3} shows additional examples of depth and normal estimation by UBotNet on real-world images. We test the network in overcast cast conditions, uneven terrain and reflective floors. UBotNet also performs well with diverse vegetation scenarios ranging from small shrubs to complex forests. We also show few results for indoor scenarios where the network performs well even though it was trained for outdoor scenarios.
Note that all the networks used for the evaluation and results discussed in this section were trained purely on OmniHorizon dataset.

% \newpage
{\small
\bibliographystyle{ieee_fullname}
\bibliography{ref}
}

\end{document}